%% file: main.tex
\documentclass{article}

\PassOptionsToPackage{numbers,sort&compress}{natbib}
 \usepackage[preprint]{neurips_2025}

\usepackage{color,soul}

\usepackage[utf8]{inputenc} 
\usepackage[T1]{fontenc}    
\usepackage{url}            
\usepackage{booktabs}       
\usepackage{amsfonts}       
\usepackage{nicefrac}       
\usepackage{microtype}      
\usepackage{xcolor}         
\usepackage{enumitem}
\usepackage[linesnumbered,ruled,vlined]{algorithm2e} 
\usepackage{amsmath}    
\usepackage{amssymb}
\usepackage{multicol}
\usepackage{makecell}   
\usepackage[most]{tcolorbox}
\usepackage{lipsum}
\usepackage{wrapfig}
\usepackage[skip=5pt]{caption} 
\input{math_commands.tex}

\usepackage{graphicx}
\usepackage{natbib}
\usepackage[hidelinks]{hyperref}
\usepackage{multirow}
\usepackage{adjustbox}
\usepackage{siunitx}

\usepackage{fontawesome5} 
\definecolor{darkpink}{rgb}{1.0, 0.07, 0.58} 
\hypersetup{
  colorlinks=true,
  urlcolor=darkpink,
}

\newcommand{\hdr}[1]{\scriptsize\textbf{#1}} 
\newcommand{\model}[1]{\textbf{#1}}
\newcounter{boxcounter}
\renewcommand{\theboxcounter}{\arabic{boxcounter}} 
\newcounter{promptcounter}
\renewcommand{\thepromptcounter}{\arabic{promptcounter}} 
\newcommand\blfootnote[1]{%
  \begingroup
  \renewcommand\thefootnote{}\footnote{#1}%
  \addtocounter{footnote}{-1}%
  \endgroup
}


\title{Bottom-up Domain-specific Superintelligence:\\ A Reliable Knowledge Graph is What We Need}

\author{%
  Bhishma Dedhia
  \qquad Yuval Kansal
  \qquad Niraj K. Jha \\[0.5pt]
  \\ Department of Electrical and Computer Engineering \\ Princeton University \\
  \texttt{\{bdedhia,\;yuvalkansal,\;jha\}@princeton.edu} \\[2pt]
  \faGlobe\enspace\href{https://kg-bottom-up-superintelligence.github.io/}{https://kg-bottom-up-superintelligence.github.io/ }\textsuperscript{\textdagger}
}

\begin{document}
\maketitle
\blfootnote{\textsuperscript{\textdagger}The project website hosts an interactive quiz built from the ICD-Bench questions; readers are encouraged to try it.}
\vspace{-2em}
\begin{abstract}
\vspace{-1em}
Language models traditionally utilized for cross-domain generalization in natural language understanding and generation have recently demonstrated task-specific reasoning through inference-time scaling. However, their top-down training approach on general text corpora is insufficient for acquiring domain-specific abstractions required for deep expertise in a particular domain. This may require a bottom-up approach that acquires deep expertise by explicitly learning to compose simple concepts of a domain into more complex ones. A knowledge graph (KG) provides such an abstraction where domain primitives are captured by head-relation-tail triples. A KG path formed by such triples captures a higher-level concept. We present a task generation pipeline that directly synthesizes tasks from the domain-specific primitives, enabling the model to explicitly acquire and compose these primitives for reasoning.  We fine-tune language models on the resultant bottom-up KG-grounded curriculum to demonstrate domain-specific superintelligence.

Although our approach is readily applicable to a wide variety of domains, we validate it in the context of medicine where reliable KGs are available.  Applying our proposed pipeline to a medical KG, we curate a dataset of 24,000 high-quality reasoning tasks paired with structured thinking traces derived from diverse medical primitives.  We fine-tune the QwQ-32B model on this bottom-up curriculum to obtain QwQ-Med-3 that takes a step towards medical superintelligence.  We also introduce an evaluation suite, ICD-Bench, to quantify domain-specific capabilities of models on reasoning tasks across 15 medical domains. Our experiments demonstrate that QwQ-Med-3 significantly outperforms state-of-the-art open-source and proprietary reasoning models on all categories of ICD-Bench. Further analysis reveals that QwQ-Med-3 utilizes acquired primitives to especially widen the performance gap on the hardest tasks in ICD-Bench. Finally, evaluation on external medical question-answer benchmarks shows that QwQ-Med-3 is able to transfer acquired expertise to improve the performance of the base model.

The industry's current approach to artificial general intelligence (AGI) centers on breadth of acquired expertise. We envision a future in which a compositional model of AGI emerges from interacting superintelligent agents, much like how the human society hierarchically acquires ever deeper expertise by combining the expertise of a group of individuals in adjacent domains or super-domains. Furthermore, since language models that are fine-tuned for superintelligence can be relatively small (e.g., 32B parameters), this bottom-up approach may also significantly cut down on training/inference energy costs.

\end{abstract}
\newpage
\begin{figure}[!hbt]
    \centering
    \includegraphics[width=\linewidth]{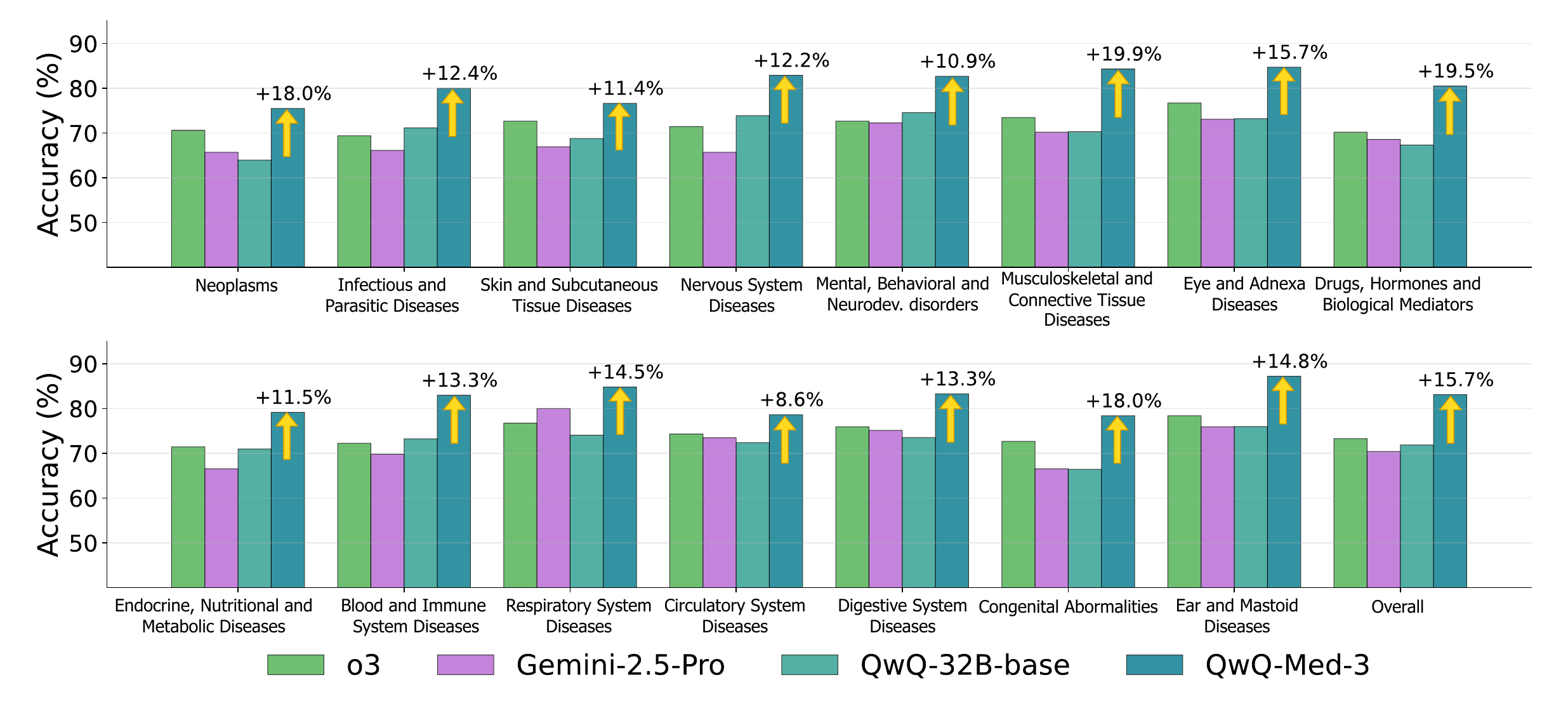}
    \vspace{-1em}
    \caption{We traverse paths on a medical KG to generate $24,000$ tasks that abstract diverse medical primitives. Our QwQ-Med-3 model, fine-tuned on these curated tasks, elicits domain-specific reasoning abilities that significantly outperform other reasoning models across medical sub-specialties. }
    \label{fig:main}
\end{figure}

\section{Introduction}

Recent advances in language modeling \cite{openai2024gpt4o,anthropic2025claude4,groeneveld2024olmo,google2024gemini15,deepseek2024v3,jiang2024mixtralexperts,yang2025qwen3technicalreport,meta2025llama4} have made a significant stride towards a cognitive system \cite{sumers2023cognitive,park2023generative} capable of performing a wide spectrum of tasks with human-like proficiency \cite{anthropic2025claudecode,openai2025codex,openai2025deepresearch,google2024gemini_deepresearch}. Yet, human-level generality may only be a waypoint on the path to advanced intelligent systems that may exceed the cognitive performance of humans: Superintelligence \cite{bostrom2014superintelligence,burns2023weak2strong}. While achieving the breadth of human cognition is one goal of advanced artificial intelligence, superintelligence might be orthogonally characterized by depth, outperforming the best human experts in specialized domains \cite{jumper2021alphafold,mankowitz2023alphadev,novikov2025alphaevolve,deepmind2025alphagenome,gottweis2025aicoscientist,merchant2023gnome,Brixi2025.02.18.638918}, like proving unsolved conjectures in number theory, developing novel kinase inhibitors for rare cancer subtypes, or discovering new ferromagnetic semiconductors that operate at room temperature. Consequently, advancing towards superintelligence might require fine-tuning general cross-domain intelligence into specialized domain-specific expertise. 

Language models (LMs) have achieved remarkable cross-domain performance in natural language understanding and generation, largely through scaled pre-training \cite{hoffmann2022trainingcomputeoptimallargelanguage,brown2020languagemodelsfewshotlearners}. Most recently, scaling inference-time compute \cite{snell2024scalingllmtesttimecompute,gao2024interpretable,muennighoff2025s1simpletesttimescaling} in pre-trained models via reinforcement learning \cite{deepseek2025r1} or post-training on high-quality data \cite{lambert2024tulu3} has been shown to elicit deeper task-specific reasoning. The emergent capacity for specialized reasoning within generalist LMs may suggest that they could serve as a foundation for scaling toward superintelligent specialists \cite{wei2022emergentabilitieslargelanguage}. However, these models are fundamentally limited by their \textit{top-down} approach to learning: they acquire general abstractions of the world through self-supervised learning on vast datasets that may predominantly capture surface-level regularities of a domain \cite{schaeffer2023emergentabilitieslargelanguage,turpin2023languagemodelsdontsay,garg2023transformerslearnincontextcase,liu2023lostmiddlelanguagemodels,qi2024safetyalignmentjusttokens,chollet2025arcagi2newchallengefrontier}. Acquiring deep expertise in a field necessitates a \textit{bottom-up} understanding, starting with axioms that capture fundamental relationships among concepts of the domain and then composing them to build upwards to a higher-order understanding \cite{Carey2009-CARTOO-3,doi:10.1126/science.1192788,lake2016buildingmachineslearnthink}. This kind of bottom-up organization is difficult to find and acquire through Internet-derived general text corpora. For example, a student builds expertise by following the pedagogical structure of a textbook, beginning with foundational chapters and gradually progressing to more advanced chapters, not merely reading encyclopedic summaries. Past pioneering works, in neurosymbolic reasoning \cite{ellis2020dreamcodergrowinggeneralizableinterpretable} and probabilistic graph inference \cite{zhang2020efficientprobabilisticlogicreasoning}, have attempted to develop hierarchical domain expertise from primitives but have failed to generalize beyond synthetic regimes. Conversely, LMs demonstrate incredible generalizability but lack grounding in structured knowledge. This motivates the central question of our work: 

\begin{center}
\textit{Can explicitly training LMs on structured domain knowledge via a bottom-up curriculum elicit the emergence (if any) of a domain-specific superintelligence?}
\end{center}
Naturally, the question then arises: How do we organize domain knowledge into a structured curriculum from which an LM can effectively learn? Knowledge graphs (KGs) \cite{Ji_2022} offer a useful scaffold for structuring knowledge that can tackle this challenge. KGs organize information as a rich graph database where nodes represent semantically meaningful entities from the domain and edges denote the relationships between them. Each edge typically captures a primitive relation in the form of a (head entity, relation, tail entity) triple. For example, \texttt{(Methane, Contains Element, Carbon)} represents the axiomatic fact that methane molecules contain carbon atoms. Edges further facilitate composite relational reasoning through the traversal of multi-hop paths along a chain of interconnected edges. For example, \texttt{(Methane, Contains Bond, C-H Bond), (C-H Bond, Is Type Of, Sigma Bond), (Sigma Bond, Has Property, Single Covalent Bond)} captures the bonding structure of methane, where \texttt{C-H} bonds are sigma bonds that possess the property of being single covalent bonds. A KG comprises many such paths whose local topology naturally induces a bottom-up curriculum, beginning with atomic relations and composing them into more complex reasoning chains.  

Our core insight is that paths in a KG can be translated into grounded natural language reasoning tasks, whose solution requires reasoning along the relational chain encoded in the paths. Training on such tasks can then enable an LM to explicitly acquire domain primitives and learn how to systematically compose them at inference. To this end, we design a task-generation pipeline that can distill high-quality reasoning tasks into a `curriculum' grounded in the KG paths. More specifically, we use a backend LM, a large language model (LLM), to map a KG path into a closed-ended question-answering (QA) task. In addition to the QA pair, we distill detailed, step-by-step thinking traces from a reasoning LLM to elucidate the relational chain on the KG path. Note that curriculum generation with a reasoning LM incurs only a one-time cost. Generating multiple such tasks across the KG and fine-tuning an LM on them yields a \textit{curriculum-tuned} model that can effectively elicit deeper reasoning (see Fig.~\ref{fig:main}) by composing acquired bottom-up KG primitives. 

While our proposed approach is domain-agnostic, given a reliable KG, we demonstrate it in the context of medicine, where such a KG is readily available.  Medical experts have meticulously curated the Unified Medical Language System (UMLS) KG \cite{Bodenreider2004TheTerminology, Yasunaga2021QA-GNN:Answering}, which offers a rich, structured repository of medical concepts and their interrelations. This makes medicine an ideal testbed for our approach: deriving a curriculum from the UMLS KG paths enables reasoning abilities beyond recalling isolated medical facts to traversing the complex web of diseases, drugs, symptoms, and underlying biological mechanisms. Models fine-tuned on UMLS-grounded curricula should be able to effectively compose learned KG primitives for reasoning across medical sub-specialties, where each domain requires deep, expert-level understanding to interpret complex clinical vignettes. To this end, we introduce ICD-Bench, an evaluation suite comprising medical QA tasks stratified across 15 categories of the International Classification of Diseases (ICD) taxonomy \cite{whoICD10}. Each ICD-Bench task is grounded in novel paths composed of domain-specific nodes from the UMLS KG, requiring the reasoning chain to correctly recall and traverse KG primitives along the path. Thus, ICD-Bench provides a reliable probe for a bottom-up domain-specific superintelligence by evaluating whether models demonstrate scalable, compositional reasoning across specialized clinical domains. 

Our experiments on ICD-Bench reveal the emergence of domain-specific reasoning in curriculum-tuned models that significantly outperform strong baselines, including proprietary and open-source reasoning models, across all 15 categories of ICD-Bench (Fig.~\ref{fig:main}). We demonstrate that performance improves with deeper and more diverse KG curricula, with curriculum depth proving especially crucial for the most challenging reasoning tasks. Fine-grained ablations further disentangle the contributions of path length, diversity, and complexity sampling, revealing that compute-optimal curricula depth distribution should adapt to task difficulty. Finally, we demonstrate that curriculum-tuned models can transfer acquired KG primitives to external medical QA benchmarks beyond the original KG. Concretely, our contributions are twofold:\vspace{-0.5em}
\begin{enumerate}[leftmargin=5pt,itemsep=1pt]
    \item We propose generating a bottom-up curriculum for LMs using a reliable domain-specific KG (Section~\ref{sec:curriculum_generation}). Our task-generation pipeline (Section~\ref{sec:data_gen_pipeline}) traverses multi-hop paths in the KG to produce natural language QA tasks grounded in domain primitives (Section~\ref{sec:qa_section}). Each QA task is paired with a high-quality thinking trace derived from its underlying KG path, enabling the construction of a training curriculum (Section~\ref{sec:data_curation}). We instantiate this framework on the expert-curated UMLS medical KG, generating $24{,}000$ QA tasks and associated thinking traces (Section~\ref{sec:curriculum_setup}). We fine-tune the off-the-shelf QwQ reasoning model on this curriculum, resulting in a curriculum-tuned model with acquired bottom-up KG primitives (Section~\ref{sec:curriculum_tuning}). 
    \item We introduce the ICD-Bench evaluation suite (Section~\ref{sec:icdbench}), comprising domain-specific QA pairs across medical sub-specialties. Our experiments demonstrate that our curriculum-tuned models significantly outperform other reasoning baselines across the ICD-Bench categories (Section~\ref{sec:domain_specific}), improve performance robustness to challenging tasks (Section~\ref{sec:robustness}), and can be optimally adapted to varying task difficulties (Section~\ref{sec:ablation}). They also demonstrate effective composition of KG primitives (Section~\ref{sec:recall_reasoning}) and transferability to other medical QA benchmarks (Section~\ref{sec:other_benchmarks}).
\end{enumerate}

\section{Background}

Next, we present background material necessary to understand the rest of the paper.

\subsection{Knowledge Graphs}

KGs represent structured knowledge as a directed graph \( \mathcal{G} \) composed of a set of node entities \( \mathcal{N} \) and their relational edges  \(\mathcal{E}\). Each edge encodes a fact that can be viewed as a triple \( (h, r, t) \), where \( h \) and \( t \) are the head and tail entities, and \( r \) is the relation linking them. For example, \texttt{(Paris, capital-of, France)} asserts that Paris is the capital of France. A path $p$ in a KG is a sequence of connected triples that forms a relational chain between two entities. A length-$N$ path is defined as:
\begin{gather}
p^N \equiv (h_0, r_1, h_1), (h_1, r_2, h_2), \dots, (h_{N-1}, r_N, h_N)
\end{gather}

Some prominent general-purpose KGs include DBpedia \cite{Auer2007DBpedia:Data}, Wikidata, and  Google's Knowledge Graph \cite{Singhal2012IntroducingStrings}, which was designed to enhance search relevance through entity linking and ranking. Enriched with ontologies that represent semantic relationships between entities (nodes) and edges, KGs facilitate complex queries and reasoning. The multi-hop paths allow the KG to capture higher-order relations between \( h_0 \) and \( h_N \). For instance, a length-3 (alternatively, a 3-hop) path could be $p^3$ = \texttt{(Marie Curie, educated-at, Sorbonne), (Sorbonne, located-in, Paris), (Paris, capital-of, France)}. This chain captures the higher-order relation that Marie Curie was educated at an institution in the capital of France. Over time, KGs have evolved from general-purpose resources to specialized graphs in critical fields; biomedical graphs, such as UMLS \cite{Bodenreider2004TheTerminology}, SemMedDB \cite{Kilicoglu2012SemMedDB:Predications}, Hetionet \cite{Himmelstein2017SystematicRepurposing}, underpin advanced applications in clinical informatics by unifying heterogeneous data into semantically rich networks.

\subsection{Unified Medical Language System (UMLS) Knowledge Graph}

In biomedicine, KGs have become indispensable for integrating disparate sources, including literature, ontologies, and clinical records to enable precision medicine and complex reasoning. At the heart of our data generation pipeline is a medical KG extracted from the comprehensive UMLS ontology. UMLS integrates multiple health and biomedical vocabularies into a unified framework by aligning synonymous terms under standardized Concept Unique Identifiers (CUIs) and linking them through a curated set of semantic relations. For instance, terms like ``myocardial infarction,'' ``heart attack,'' and ``MI'' are all mapped to the same CUI. Past pioneering work \cite{Bodenreider2004TheTerminology} has constructed and filtered a KG from UMLS by representing each CUI as a node and the semantic relationships, such as ``treats,'' ``causes,'' or ``is a subtype of,'' as directed edges between nodes. Thereafter, researchers \cite{Yasunaga2021QA-GNN:Answering} combined the disease sub-part of the  UMLS KG with DrugBank \cite{Wishart2018DrugBank} to create an expansive drug-and-disease database. We traverse the paths of this constructed KG to generate a curriculum of medical relations ordered by path length. Simple 1-hop paths correspond to simple medical factoids such as Aspirin $\rightarrow$ may-treat $\rightarrow$ Myocardial Infarction. In contrast, more complex multi-hop paths can support clinically meaningful vignettes. For example, the path
Diabetes Mellitus $\rightarrow$ predisposes to $\rightarrow$ Kidney Diseases $\rightarrow$ causes $\rightarrow$ Anemia
captures the reasoning chain that diabetes may lead to kidney disease, which in turn can cause anemia due to impaired erythropoietin production.

\subsection{The International Classification of Diseases (ICD)}
\label{sec:icd}
Our work investigates the emergence of domain-specific expertise in fine-tuned LMs, specifically as it manifests in medical reasoning. However, medicine is a broad field, encompassing a complex topology of specialized subfields, each dedicated to the diagnosis and treatment of distinct categories of health conditions. To be truly useful, a superintelligent medical specialist must not only grasp the broad scope of medicine but, more importantly, demonstrate the ability to reason effectively within narrow, highly specialized domains. It is, therefore, necessary to evaluate and benchmark the capabilities of medical specialists across these specialized domains. The ICD \cite{whoICD10}, a globally recognized taxonomy for recording, reporting, and analyzing health conditions, provides a natural structure for this effort (Fig.~\ref{fig:icd} top). We leverage the ICD framework to design a benchmark called ICD-Bench (Section~\ref{sec:icdbench}), which systematically evaluates domain-specific medical reasoning along distinct axes of ICD disease types.

\section{Bottom-Up Curriculum Generation For Language Models}
\label{sec:curriculum_generation}
\textbf{Overview}: First, we use a KG as a scaffold for generating closed-ended tasks that are grounded in the structured entities of the KG. To this end, we traverse local paths of a KG to construct grounded tasks in the form of \textbf{(question, answer)} pairs. We then extract high-quality thinking traces for each QA pair from a reasoning language model grounded in the traversed path. During training, we fine-tune an LM on a curated curriculum of \textbf{(question, thinking trace, answer)} triplets generated by this pipeline. At inference time, we scale inference-time compute on the curriculum-tuned model.

\subsection{The Task-Generation Pipeline}
\label{sec:data_gen_pipeline}
The ultimate goal of our task-generation pipeline is to curate reasoning tasks using KGs, guided by three core design principles: \vspace{-0.5em}
\begin{itemize}[leftmargin=*,itemsep=0.5pt]
\item \textbf{Closed-endedness}: Each generated task should have a distinct correct answer with the reasoning traceable to grounded paths in the KG. 
\item \textbf{Steerable Complexity}: The pipeline should facilitate reliable control over the depth of reasoning required to solve the generated tasks. 
\item \textbf{Diversity}: The pipeline should ensure that traced paths cover the entire KG instead of being concentrated on a few nodes.
\end{itemize}

\subsubsection{Generating Grounded Question-Answering Tasks Using a KG}
\label{sec:qa_section}

\begin{figure}
    \centering
    \includegraphics[width=\linewidth]{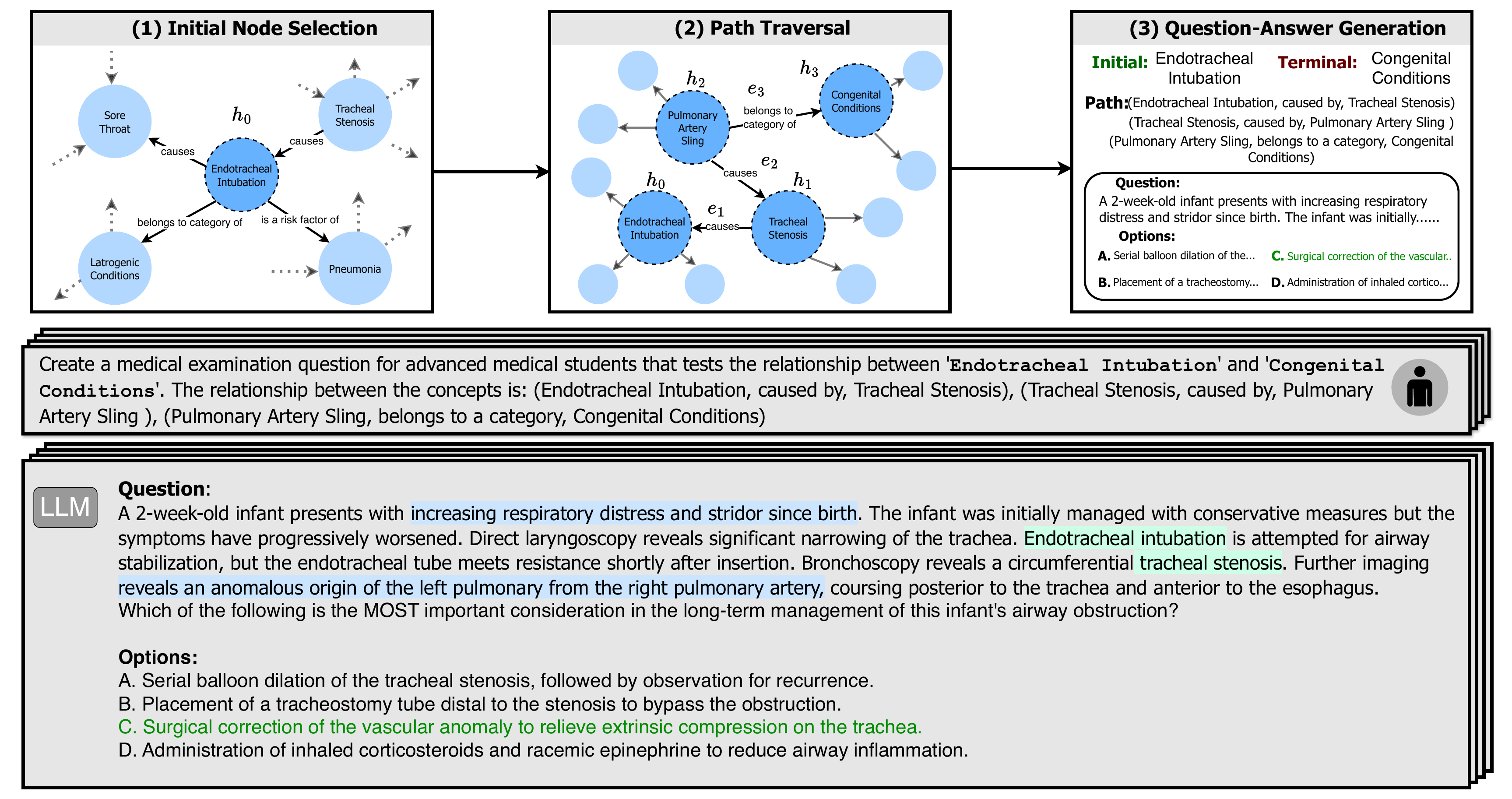}
    \caption{\textbf{Generating QA tasks from a KG path.} We explore KG paths to derive QA pairs grounded in the KG. We choose a KG path by starting from an initial node (left) and iteratively sampling (relation, entity) pairs from the current node's neighbors to obtain an $N$-hop path $p^N$ (middle). The sampled path is mapped to a natural language QA task by prompting a backend LLM (right). The bottom of the diagram shows an example of a generated QA pair, where text highlighted in green indicates entities revealed in the question and blue highlights indicate latent entities. Effectively solving the QA task requires recalling latent entities and reasoning along the KG path to reach the correct solution.}
    \label{fig:qa_generation}
\end{figure}

Each data point from our pipeline is instantiated as a multiple-choice question sourced from the KG. This QA format encourages reasoning as models must generate rationales for both identifying the correct answer and eliminating distractor options. The generation process, illustrated in Fig.~\ref{fig:qa_generation} (top), comprises three main stages:

\textbf{(1) Initial Node Selection:} We begin by selecting an initial concept node $h_0$ from the KG.

\textbf{(2) Path Traversal:} From \(h_0\), we sample a multi-hop path of length $N$ on the KG in $N$ steps. At each step \(t\), we consider the set of all outgoing (relation, neighbor) pairs from the current node \(h_{t-1}\), exclude any neighboring nodes already visited, and sample one pair \((e_t, h_t)\) uniformly. This simultaneously selects the relation \(e_t\) and the next node \(h_t\) in one draw. More formally, given $h_0$ and path-length $N$:
\[
\begin{aligned}
&\textbf{Initialize Path:} \quad p^0 = \emptyset  \\[1pt]
&\textbf{For } t = 1 \text{ to } N: \\[2pt]
&\quad \textbf{Gather Candidates:} \quad 
\mathcal{C}_t = \left\{(e, v) \mid (h_{t-1}, e, v) \in \text{Neighbors}(h_{t-
1}),\ v \notin \{h_0, \dots, h_{t-1}\} \right\} \\[2pt]
&\quad \textbf{Sample Next Hop:} \quad 
(e_t, h_t) \sim \mathrm{Uniform}(\mathcal{C}_t) \\[2pt]
&\quad \textbf{Extend Path:} \quad 
p^{t+1} \leftarrow p^t \cup  (h_{t-1}, e_t, h_t)
\end{aligned}
\]

Here, \(\mathcal{C}_t\) collects all valid outgoing pairs \((e,v)\) from the current node \(h_{t-1}\), excluding those whose target \(v\) has already appeared in \(\{h_0,\dots,h_{t-1}\}\).  After \(N\) hops, the complete path is  $(h_0,r_1,h_1),\cdots,(h_{N-1},r_N,h_N)$. 

\textbf{(3) Question-Answer Generation:} Our pipeline transforms each sampled KG path into a question-answering task by leveraging a backend LLM. Specifically, we design a template prompt that tasks the model with constructing a vignette (a clinical one in the case of medical superintelligence) and posing a multiple-choice question whose resolution depends on traversing the entire path. The template\vspace{-0.5em}
\begin{enumerate}[leftmargin=10pt]
\item instructs the model to formulate a question that links the initial node $h_0$ to the terminal node $h_{N}$, 
\item provides the complete path as context to ensure factual grounding, and \item enforces a correct answer with other plausible but false options. 
\end{enumerate}

The transformation can be formalized as follows:
\begin{equation*}
\text{Question, Answer} = LM\big(\;\mathcal{T}\bigl(\;\text{source}=h_0,\;\text{target}=h_N\;,\text{context}=p^N\bigr)\big)
\end{equation*}

Here, $\mathcal{T}(.)$ denotes the template-based prompt (see Appendix~\ref{app:qa_generation_template}) and LM(.) denotes sampling from an LLM. We use the \texttt{Gemini 2.0 Flash} model \cite{deepmind_gemini2.5flash_2025} to generate the QA pairs. Fig.~\ref{fig:qa_generation} (bottom) shows a QA example along with its path context. Returning to the aforementioned design principles, the QA format naturally endows closed-endedness since the model must provide a single correct answer, the source node and the subsequently sampled path provide a natural control over the diversity of the QA task, while the path length allows for steering complexity.

\subsubsection{Curriculum Curation}
\label{sec:data_curation}
\begin{figure}[!tbh]
    \centering
    \includegraphics[width=\linewidth]{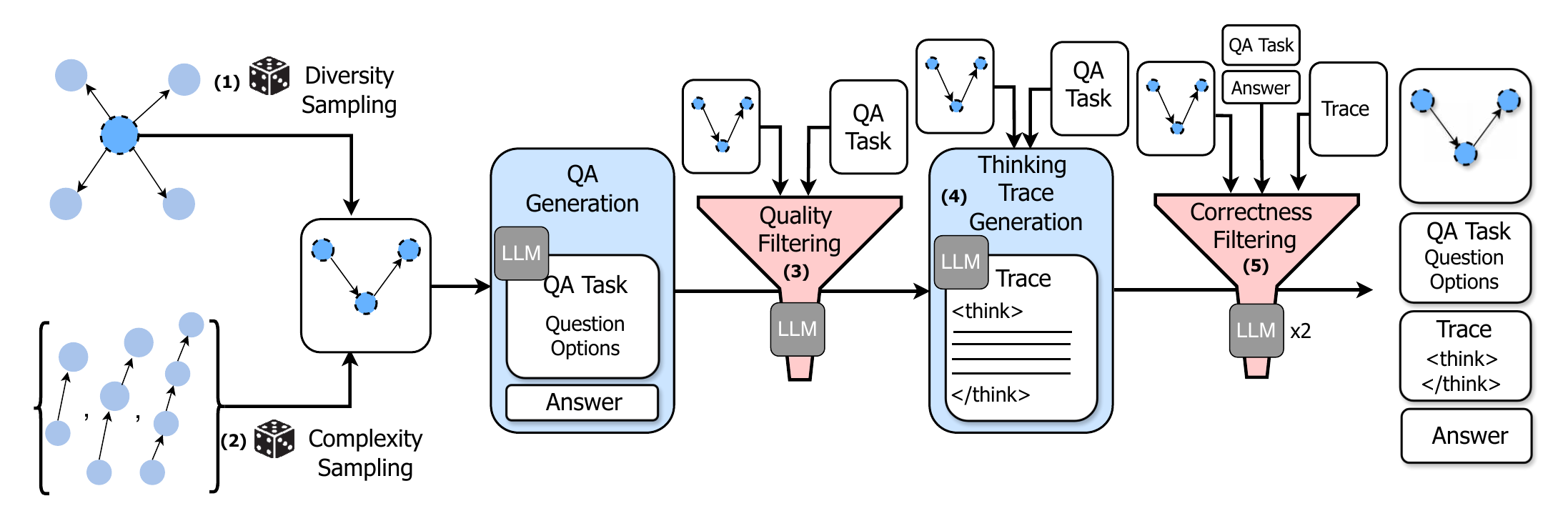}
    \captionsetup{belowskip=0pt,aboveskip=4pt}
    \caption{\textbf{Overview of our curriculum curation pipeline.} Starting from KG-derived paths, we sample for node diversity and path complexity (Steps 1–-2), followed by quality filtering of generated QA pairs (Step 3). We then generate thinking traces using a strong reasoning LLM grounded in the KG path (Step 4) and, finally, perform  correctness filtering using two independent grader LLMs to ensure factual trace alignment to the KG path and answer validity (Step 5).}
    \label{fig:pipeline}
\end{figure}

We use the transformation method to leverage the KG path to QA pair to assemble a training curriculum of high-quality QA pairs. The dataset is carefully curated for \textit{diversity} of source nodes across the KG, \textit{complexity} via enforcing different-length hops, and \textit{quality and correctness} by introducing filtering heuristics for QA coherence. The pipeline, illustrated in Fig.~\ref{fig:pipeline}, proceeds via the following steps.

\textbf{(1) Diversity Sampling:} To ensure that our sampled paths provide broad coverage of the KG and avoid clustering around a small subset of highly-connected nodes, we enforce diversity while selecting the source node. During the data curation process, we maintain a running selection frequency with which nodes are sampled on paths. Then we sample the source node based on the inverse of the selection frequency, ensuring that unsampled nodes or fewer-sampled nodes are sampled more. Let $f_i$ denote the sample frequency of node $i$ in the set of generated QA pairs. Then node $i$ is sampled as the source node with probability:
\begin{equation*}
    p_i = \frac{w_i}{Z}\;;\quad \text{where}\; w_i = \frac{1}{f_i+\epsilon},\;Z = \sum^{|\mathcal{N}|}_{i=1} w_i
\end{equation*}
$\epsilon\;(=1)$ is a small constant that prevents division by zero for unsampled nodes.

\textbf{(2) Complexity Sampling:} To induce a graded notion of reasoning difficulty, we uniformly sample KG paths of varying lengths from $\{1,\cdots,N\}$ instead of always selecting the longest possible paths. This introduces a natural curriculum where shorter paths typically yield recall-based or single-hop queries, while longer paths require multi-hop, compositional reasoning. By training on a range of path complexities, the model develops balanced reasoning skills and avoids overfitting to long, potentially noisy chains, thereby improving robustness and generalization. We ablate and show the effect of sampled KG depth on reasoning performance in Section~\ref{sec:ablation}.

\textbf{(3) Quality Filtering:} To ensure high-quality QA pairs, we implement a multi-stage filtering process. We first discard generations with application programming interface (API) call failures, incomplete responses, or distracting artifacts, such as ASCII strings or code blocks. Next, we enforce strict adherence to a predefined QA template. Each question must be phrased as a vignette grounded in a KG path, followed by one correct answer and three plausible distractors, with consistent formatting (e.g., option labels like `A.'). Finally, we eliminate QA pairs with low-quality distractors, such as near-duplicates or distractors that closely resemble the correct answer, to preserve the discriminative integrity of each question.

\textbf{(4) Thinking Trace Generation:} After QA pairs pass quality filters, we distill high-quality thinking traces in natural language from the underlying KG paths. For each retained question, we prompt a strong reasoning model with the vignette and options from the QA pair with the full KG path as context, as follows:
\begin{equation*}
\text{Thinking Trace} = LM\big(\text{question = (Vignette, Options)},\text{context}=p^N\big)
\end{equation*}
More specifically, we distill the traces from the \texttt{Gemini-2.5-Pro} model \cite{deepmind_gemini2.5pro_2025}, which has demonstrated state-of-the-art reasoning capabilities. We instruct the model to reason through the question to infer the correct answer, referencing the KG path (see Appendix~\ref{app:task_generation_pipleline}, Prompt~\ref{prompt_template_trace}). By anchoring the reasoning trace to the KG path, we produce structured rationales grounded in the KG that ensure strong relational supervision for fine-tuning. 

\textbf{(5) Correctness Filtering:} Despite grounding questions in explicit KG paths, errors can arise due to ambiguous phrasing, inconclusive evidence along the KG path, or LLM hallucinations. To address this, we perform a final correctness check to ensure that each QA item (question, thinking trace, answer) is unambiguously interpretable based on the provided path and that the thinking trace faithfully follows the path to arrive at the correct answer without hallucinations. We organize the complete context of the QA item under a template prompt and task an LLM grader to verify correctness. We specifically instruct the grader model (see Appendix~\ref{app:task_generation_pipleline}, Prompt~\ref{prompt_template_correctness}) to evaluate whether (a) the correct answer follows from the vignette and the KG path, and (b) every claim in the thinking trace is supported by the KG path, without hallucinations. The grader outputs a binary verdict. To guard against idiosyncratic failures of any single model, we enforce a two-factor agreement using two grader models -- \texttt{Gemini 2.0 Flash} and \texttt{Qwen 2.5-72B} \cite{qwen2025qwen25technicalreport}. We retain QA items only if both independent grader models verify correctness, ensuring robustness through cross-model consistency.

We iteratively repeat these steps until we have a user-defined size of high-quality QA items. The task-generation pipeline has been summarized in 
Appendix~\ref{app:task_generation_pipleline} Algorithm~\ref{alg:curriculum}.

\subsection{Curriculum Tuning and Inference}
\label{sec:curriculum_tuning}
\textbf{Curriculum Tuning:} We use our curated dataset, specifically reasoning traces derived from KG paths, to perform supervised fine-tuning (SFT) of off-the-shelf LMs via the next token prediction objective. Prior to training, we map each (question, thinking trace, answer) datapoint to a chat template, with the thinking trace inserted between special \texttt{<think>} and \texttt{</think>} delimiters to signal the beginning and end of the thinking process, respectively. We refer to the resulting fine-tuned models as \textit{curriculum-tuned} models, since they have been explicitly trained to acquire a structured reasoning curriculum grounded in KG primitives.

\textbf{Inference:} At inference time, we scale compute on our curriculum-tuned models by expanding the generated thinking trace \cite{snell2024scalingllmtesttimecompute}, either by generating multiple traces in parallel \cite{wei2022emergentabilitieslargelanguage,wang2023selfconsistencyimproveschainthought} or by extending individual traces through iterative refinement \cite{muennighoff2025s1simpletesttimescaling}. Specifically:
\begin{itemize}[leftmargin=*, itemsep=1pt]
\item \textbf{Parallel Scaling:} We generate $n$ independent thinking traces for each test question in parallel. Each instance produces a complete trace (delimited by \texttt{<think>} and \texttt{</think>}), followed by an answer. The final prediction is obtained via majority voting across the $n$ outputs.
\item \textbf{Iterative Refinement:} We also allocate additional compute to trace refinement, encouraging the model to re-evaluate its reasoning. Following prior work \cite{muennighoff2025s1simpletesttimescaling}, we intervene in the decoding process by replacing the end-of-thinking delimiter \texttt{</think>} with prompts like \texttt{“hmm, let's double check”}, prompting the model to continue its thought process before finalizing an answer.
\end{itemize}

\begin{figure}
    \centering
    \includegraphics[width=\linewidth]{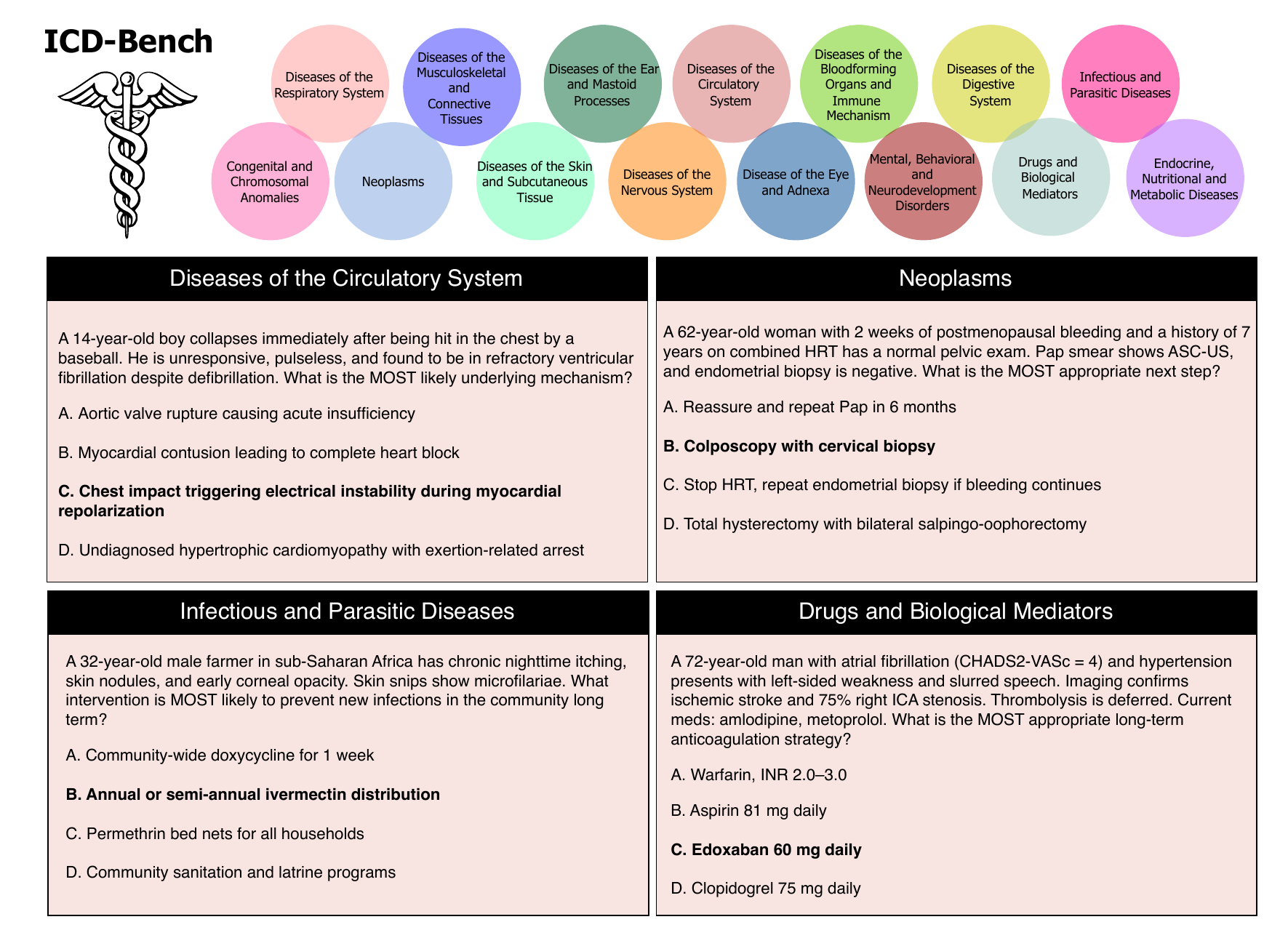}
    \captionsetup{belowskip=0pt,aboveskip=4pt}
    \caption{\textbf{ICD-Bench evaluation suite} Top: The 15 medical sub-specialties derived from the ICD-10 taxonomy, each corresponding to a distinct category in the benchmark. Each node in the UMLS KG is mapped to one or more of these categories to guide domain-specific QA generation. Bottom: Sample QA items drawn from different ICD categories, illustrating the diversity of the benchmark in medical reasoning tasks, from treatment selection and diagnostic evaluation to mechanistic and public health interventions across disease types.}
    \label{fig:icd}
\end{figure}

 \section{ICD-Bench Evaluation Suite}
\label{sec:icdbench}

To rigorously evaluate the emergence of domain-specific reasoning capabilities in our curriculum-tuned models, we introduce \textbf{ICD-Bench}, a targeted QA benchmark aligned with the ICD taxonomy (see Section~\ref{sec:icd}). The benchmark is constructed using structured knowledge embedded in the UMLS KG and is designed to evaluate models on domain-specific tasks spanning medical sub-specialties. We begin by describing its construction.

\subsection{ICD-Bench Construction Procedure}
\textbf{(1) Aligning the KG to the ICD-10 Taxonomy:} To enable fine-grained control over domain-specific QA generation, we map each node in the UMLS KG to one or more of the 15 ICD categories illustrated in Fig.~\ref{fig:icd} (top). This mapping is performed by an LLM classifier, which assigns categories only to nodes with a strong affinity. 

\textbf{(2) Question Generation:} We use the stratified KG to generate hop-controlled QA items per category, using our QA generation method (Section~\ref{sec:qa_section}), as follows: 
\begin{enumerate}[leftmargin=*,itemsep=3pt]
    \item\textbf{Select Category:} Choose category $C$ from the ICD-10 taxonomy.
    \item\textbf{Sample Source Node:} Sample a source node $h^0$ belonging to category $C$.
    \item\textbf{Select Path Complexity:} Choose path complexity $N$.
    \item\textbf{Generate QA:} Traverse a length-$N$ KG path $p^N$ beginning at $h^0$ and generate a QA pair.
\end{enumerate}

\textbf{(3) Quality and Correctness Filtering:} Finally, we subject the generated QA pairs to quality and correctness checks using the steps outlined in Section~\ref{sec:data_curation}.

\subsection{ICD-Bench Composition} 

The final ICD-Bench evaluation suite comprises 3,675 high-quality QA items, systematically generated through the controlled pipeline described above. These items are evenly distributed across the $15$ ICD-10 categories, enabling a balanced assessment of reasoning capabilities across diverse medical domains. Each category contributes 245 QA items, stratified by $100$ questions derived from two-hop KG paths, $100$ from three-hop paths, $30$ from four-hop paths, and $15$ from five-hop paths. Tasks from on one-hop paths are omitted to minimize bias towards those that require simple recall or memorization, rather than deeper reasoning. This structure ensures that ICD-Bench probes both the breadth of domain-specific knowledge and the depth of compositional reasoning needed to navigate each domain. Fig.~\ref{fig:icd} (bottom) shows representative QA examples drawn from distinct ICD categories, spanning simple entity-relational queries and composite chains involving treatments, diagnostics, and etiological factors.

\section{Experiment Setup}
\label{results}

Next, we define the experimental setup.

\subsection{Training Curriculum Curation and Decontamination Setup}
\label{sec:curriculum_setup}
\textbf{Curriculum Curation:} We leverage our proposed task-generation pipeline (Section~\ref{sec:data_gen_pipeline}) to curate a training curriculum of $24{,}000$ QA tasks along with their thinking traces on the UMLS KG. The generated tasks span diverse medical entities and relations on the KG, and are distributed uniformly over multi-hop lengths $\in \{1,\;2,\;3\}$. We restrict hop lengths to $N\leq3$ based on an empirical observation that paths longer than three hops often traverse semantically unmeaningful and weak relations, diminishing the coherence and correctness of the resulting questions. Moreover, we reserve the small subset of semantically meaningful long-range paths ($N\geq4$) for evaluation via ICD-Bench. 

\begin{figure}[!bth]
    \centering
    \includegraphics[width=0.85\linewidth]{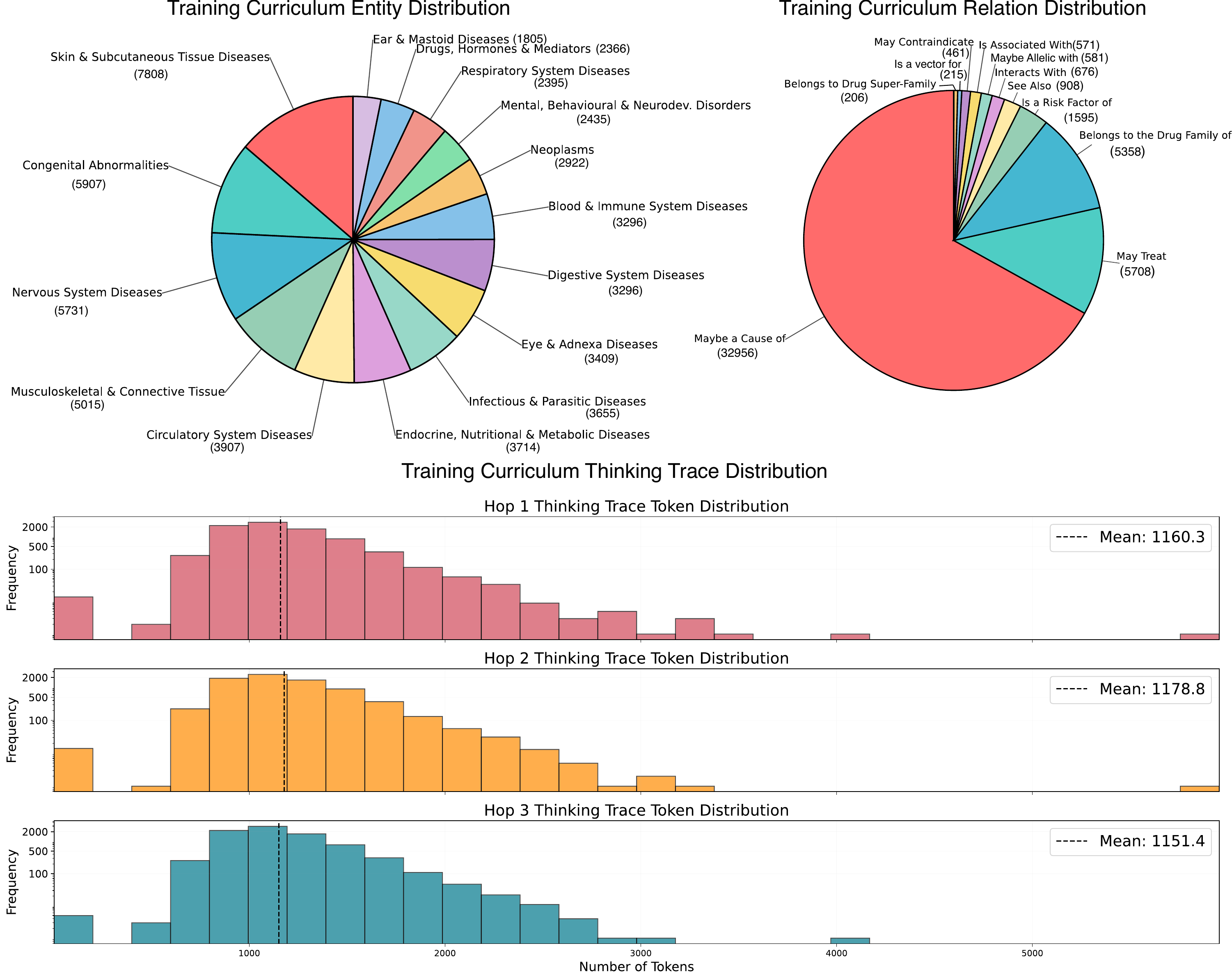}
    \captionsetup{belowskip=0pt,aboveskip=4pt}
    \caption{\textbf{Distributional statistics of the curated training curriculum.} The dataset spans 24,000 QA items grounded in UMLS KG paths. Left: Breakdown of entities in the sampled KG paths, across the ICD categories they belong to. Right: Distribution of relation types along sampled KG paths, spanning different semantic relations. Bottom: Hop-wise distribution of thinking trace lengths across the dataset, reflecting variance in reasoning complexity.}
    \label{fig:train_dist}
\end{figure}

\textbf{Decontamination:} We also perform a two-fold decontamination of our generated tasks prior to their inclusion in the training data. First, we exclude any QA pair where the underlying KG path \textit{exactly} traverses a KG path of any ICD-Bench QA pair. This prevents memorized KG paths from contaminating evaluations. However, we allow \textit{partial} path overlaps, as our goal is to allow models to learn and generalize from individual KG primitives. Therefore, we additionally adopt an $18$-gram threshold for the text overlap filter to eliminate QA pairs that are highly similar to those in ICD-Bench, filtering out close matches while preserving distinct yet conceptually related questions. 

Fig.~\ref{fig:train_dist} displays the breakdown of category entities (left) and  relations (right) in the KG curriculum, and the distribution of the thinking trace lengths across multi-hop paths (bottom). Examples of generated tasks and token breakdown of the curriculum are provided in Appendix~\ref{app:UMLS_curriculum}. 
 
\subsection{Curriculum-Tuning Setup} 
Our work investigates how domain-specific reasoning abilities can emerge in a general-purpose model through fine-tuning on structured curricula derived from a domain KG. We adopt the open-source QwQ-32B LM \cite{yang2025qwq32b} as our base, leveraging its strong reasoning foundations acquired via large-scale reinforcement learning. To study how a curriculum affects generalization, we construct three training datasets that progressively incorporate tasks from deeper KG paths and fine-tune the base model on each dataset under a fixed floating-point operations (FLOPs) budget. This yields three fine-tuned models:
\begin{itemize}[leftmargin=1em,itemsep=0.5em]
\item \textbf{QwQ-Med-1:} Fine-tuned on $8{,}000$ tasks derived from single-hop KG paths for $24$ epochs.
\item \textbf{QwQ-Med-2:} Trained on $16{,}000$ tasks that combine one-hop and two-hop paths for $12$ epochs.
\item \textbf{QwQ-Med-3:} Extended to include three-hop paths, totaling $24{,}000$ tasks over $8$ epochs.
\end{itemize}
From a pedagogical standpoint, each successive model is trained on a curriculum that grows both deeper and wider with respect to the underlying KG. Depth increases through the inclusion of longer multi-hop reasoning chains, while breadth expands as the model is exposed to a more diverse set of KG entities and relational contexts. All models are fine-tuned using low rank adapters (LoRA) \cite{Hu2021LoRA:Models} with rank $16$ on $8$ H100 NVIDIA GPUs, with each run taking approximately $20$ hours. The complete SFT-through-LoRA setup is outlined in Appendix \ref{app:training_details}.

\section{Experiments}
\label{experiments}

Next, we present our experiments.

\subsection{Understanding Inference-Time Scaling Behavior of Curriculum-Tuned Models}
\begin{figure}
    \centering
    \includegraphics[width=\linewidth]{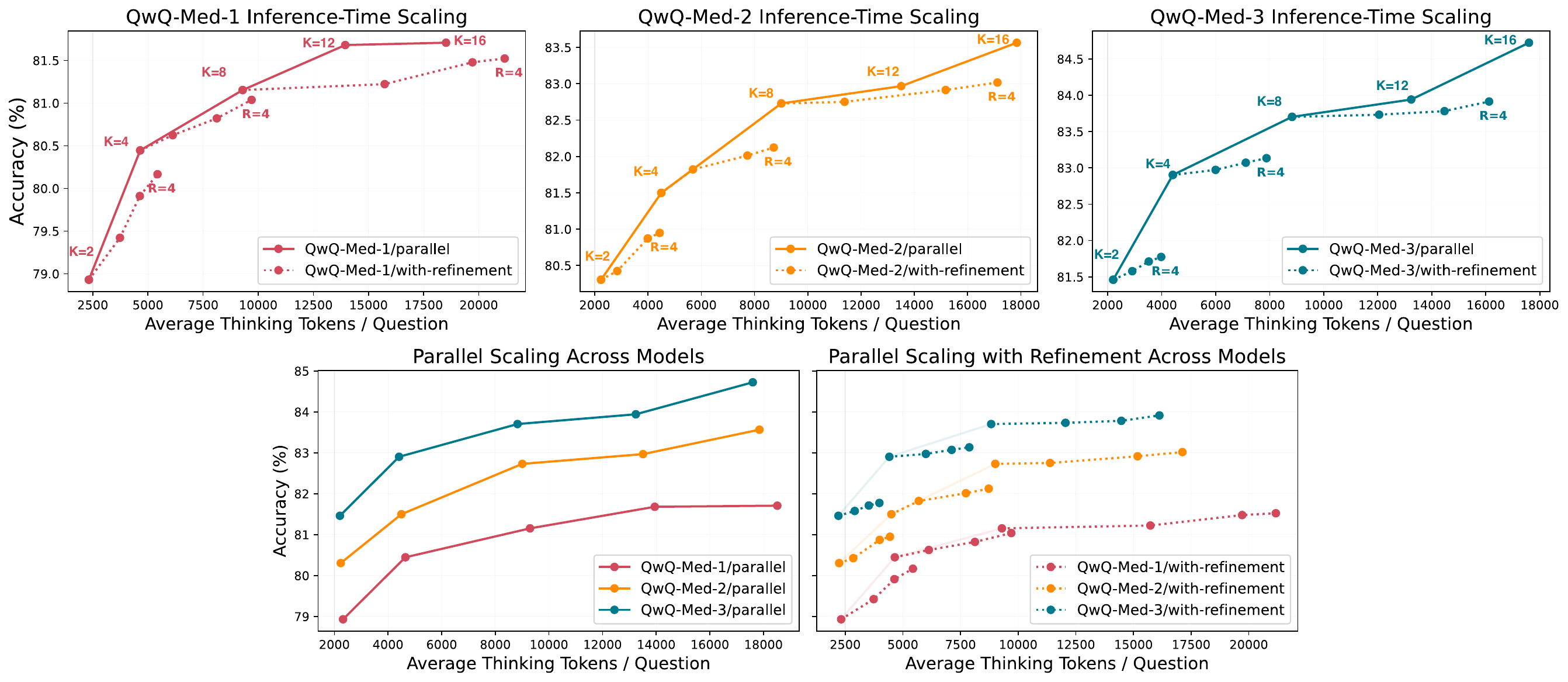}
    \captionsetup{belowskip=0pt,aboveskip=4pt}
    \caption{\textbf{Inference-time scaling curves for curriculum-tuned models on ICD-Bench.} Top row: Each plot displays per-model curves where solid lines denote pure parallel scaling and dotted lines denote iterative refinement augmentation. Deeper curriculum models (QwQ-Med-2, QwQ-Med-3) benefit more from parallel scaling, while QwQ-Med-1 remains amenable to refinement. Bottom row: Each plot shows a comparison of per-scaling technique curves. QwQ-Med-3 trained on the entire generated curriculum demonstrates compute-optimality over models trained on partial curricula. Bootstrapped confidence intervals over $500$ samples were $<0.75\%$.}
    \label{fig:ttc}
\end{figure}

\textbf{(S1) Setup:} We evaluate our curriculum-tuned models on ICD-Bench by extending inference-time compute using the parallel and iterative refinement strategies outlined in Section~\ref{sec:curriculum_tuning}. For parallel inference, we vary the number of concurrent reasoning streams with $K \in \{2, 4, 8, 12, 16\}$, setting decoding temperature to $0.6$. For iterative refinement, we further augment each parallel stream with $R=4$ refinement steps, evaluating this setting for $K \in \{2, 4, 8\}$. We evaluate each model-compute configuration on the full set of $3{,}675$ ICD-Bench questions across $15$ medical categories. We report the overall accuracy alongside the average number of thinking tokens consumed per question. The inference-time scaling curves for our models are shown in Fig.~\ref{fig:ttc}. We observe that:

\textbf{(O1.1) Parallel scaling outperforms iterative refinement with increasing curriculum.}
 QwQ-Med-2 and QwQ-Med-3 exhibit steeper gains from parallel scaling (solid lines) while refinement (dotted lines) saturates, unlike QwQ-Med-1, where improvements from parallelism asymptote and converge with refinement (top row, Fig.~\ref{fig:ttc}). This dichotomy indicates that deeper curriculum models, having acquired structured and diverse KG primitives, benefit more from exploring multiple reasoning paths in parallel. In addition, unlike math and coding tasks, where refinement helps due to verifiable intermediate steps, medical diagnosis hinges on early differential diagnosis, making parallel sampling more impactful for curriculum-tuned models. 

\textbf{(O1.2) Structured curriculum-tuning enables inference-time compute optimality.} With an increasingly difficult curriculum, our fine-tuned models achieve higher accuracy at lower inference budget, as reflected by leftward shifts in the scaling curves (bottom row, Fig.~\ref{fig:ttc}). Despite being fine-tuned on an equal training FLOPs budget, models learn to allocate inference budget more effectively. We posit that exposure to deeper multi-hop chains and broader KG coverage allows models to acquire and reuse reasoning derived from KG primitives, which enables them to converge to accurate answers with less iterative search or brute-force sampling at test time.

In the rest of the paper, we use parallel inference-time scaling unless otherwise mentioned.

\label{sec:domain_specific}

\subsection{Domain-Specific Reasoning Emerges from Curriculum-Tuned Models}

\textbf{(S2) Setup:} We gain insight into the domain-specific capabilities acquired through curriculum tuning by evaluating our models on the category-specific branches of ICD-Bench. Each subset comprises 245 medical QA examples focused on a specific disease type, allowing us to disentangle performance gains along distinct clinical axes.

\textbf{Baselines:} We compare our curriculum-tuned models against four baselines: (1) the QwQ-32B base model, which serves as our reference general-purpose reasoning model, (2) DeepSeek-R1-Distilled Qwen \cite{deepseek2025r1}, another strong open-source reasoning model distilled from Deepseek-R1, and (3) two proprietary state-of-the-art reasoning models, o3 \cite{openai_o3_o4mini_2025} and Gemini-2.5-Pro \cite{deepmind_gemini2.5pro_2025}, known for strong domain generalization and competitive benchmark performance. We perform inference-time scaling on the open-source models while reporting pass$@1$ accuracy (of the first generated solution) for the proprietary models. 

\begin{figure}[!hbt]
    \centering
    \includegraphics[width=\linewidth]{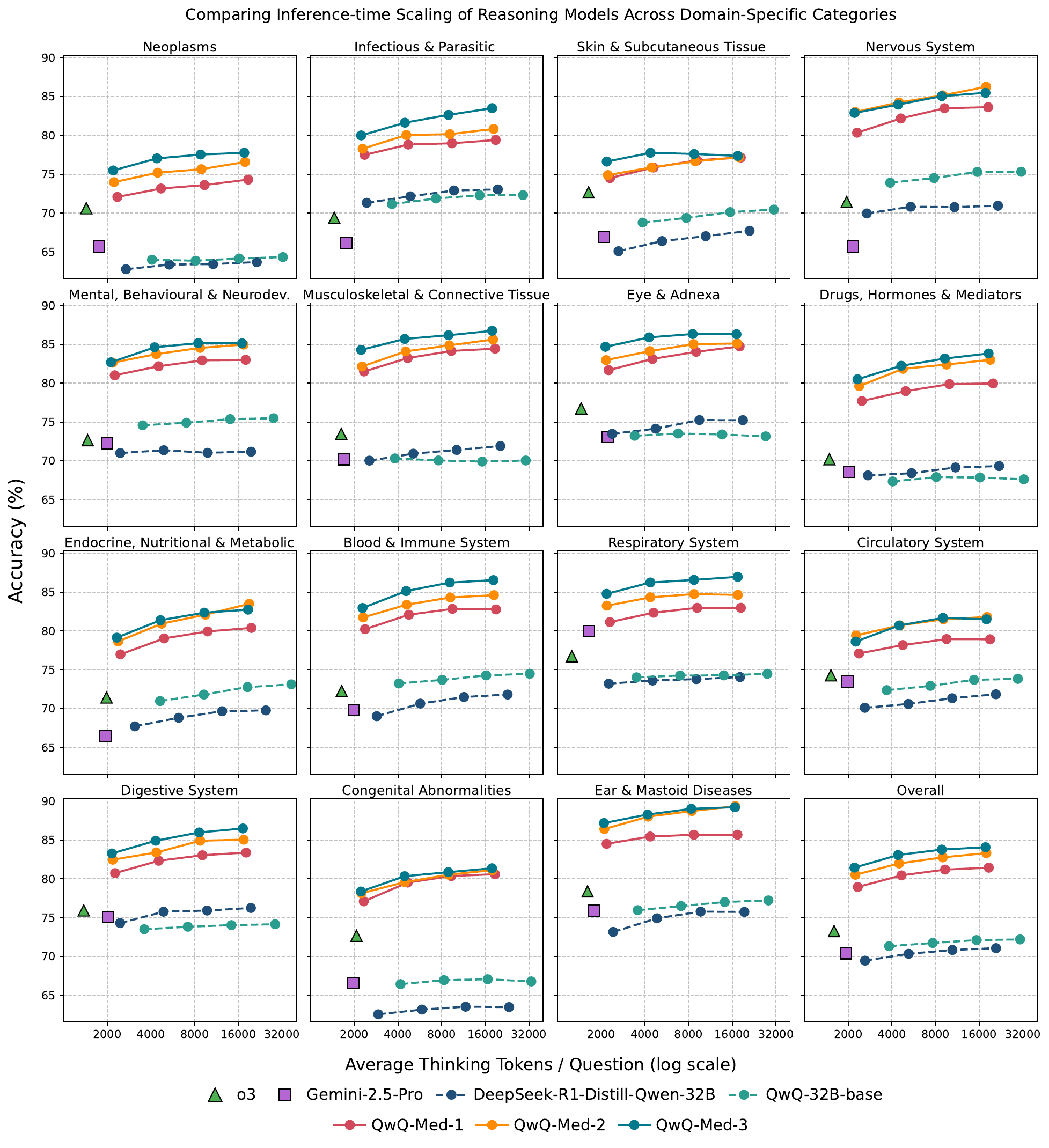}
    \captionsetup{belowskip=0pt,aboveskip=4pt}
    \caption{\textbf{Domain-specific performance of our curriculum-tuned models across ICD-Bench categories.} Curriculum-tuned models significantly outperform proprietary and open-source reasoning baselines across medical sub-specialities. Moreover, curriculum-tuned models demonstrate improved inference-time scaling, with greater curriculum (QwQ-Med-3) enabling compute-optimality across categories. Bootstrapped confidence intervals over $500$ samples were $<0.75\%$.}
    \label{fig:category_wise_scaling}
\end{figure}

In Fig. \ref{fig:main}, we present the performance of our model relative to proprietary baselines across ICD-Bench categories. Fig. \ref{fig:category_wise_scaling} shows scaling results of reasoning models across ICD-Bench categories. Our major takeaways are: 

\textbf{(O2.1) Curriculum-tuned models significantly outperform other reasoning models.} Our models consistently outperform all open-source baselines across inference budgets by 10-20\%. Moreover, open-source reasoning models tend to plateau early with increasing compute, whereas our curriculum-tuned models demonstrate a better utilization of inference-time budget. Strikingly, our models also outperform o3 and Gemini-2.5-Pro, despite their massive model size and training on web-scale data. While proprietary models are competitive in highly prevalent disease cases, like neoplasms, circulatory, and respiratory conditions, that are more frequently represented in text corpora, our curriculum-tuned models show crucial improvements in less prevalent categories, like congenital abnormalities and nervous system disorders. 

\textbf{(O2.2) Expanding the curriculum improves domain-specific reasoning.} Across most ICD-Bench categories, QwQ-Med-3 outperforms both QwQ-Med-2 and QwQ-Med-1, highlighting the cumulative benefit of scaling curriculum depth and diversity. In some categories, QwQ-Med-2 and QwQ-Med-3 perform similarly, which may reflect early saturation from acquiring densely linked support KG paths that already capture the central reasoning primitives needed for that disease type. 

\textbf{Qualitative.} In Examples~\ref{qual_1} and~\ref{qual_2}, we present sample responses from our QwQ-Med-3 model. Each entity and relation from the underlying KG path is color-coded distinctly. Corresponding segments in the model's response that recall these entities or trace the reasoning along the relations are highlighted in the same color. This alignment illustrates that curriculum-tuned models can recall acquired KG primitives and coherently compose them during reasoning. Additional outputs and comparison to the base model are provided in Appendix~\ref{app:qualitative}. 

\subsection{Curriculum-Tuned Models Improve Robustness to Task Difficulty} 
\label{sec:robustness}
Real-world clinical reasoning tasks, such as those in ICD-Bench, exhibit a spectrum of complexity, from direct factual recall to implicit, multi-step inference.  In this section, we probe whether our curriculum-tuned models exhibit improvements across this full difficulty range, specifically whether the acquisition of KG primitives enables them to reliably reason on hard tasks.

\textbf{Task Difficulty Estimation:} To estimate difficulty, we use the base QwQ model as a proxy evaluator. For each task, we compute its pass@$1$ rate, the fraction of times the model produces a correct answer across $16$ independently sampled generations. This score serves as a difficulty heuristic, with lower rates signifying harder tasks. The resulting distribution is bimodal, with a dominant mass near $100\%$ and a secondary mode at lower success rates. Subsequently, we partition the tasks into five difficulty bins based on this empirical distribution, capturing a fine-grained spectrum of task hardness (see Appendix~\ref{app:additional_robustness} for difficulty distribution and difficulty bin cutoffs).

\begin{figure}[!tbh]
    \centering
    \includegraphics[width=\linewidth]{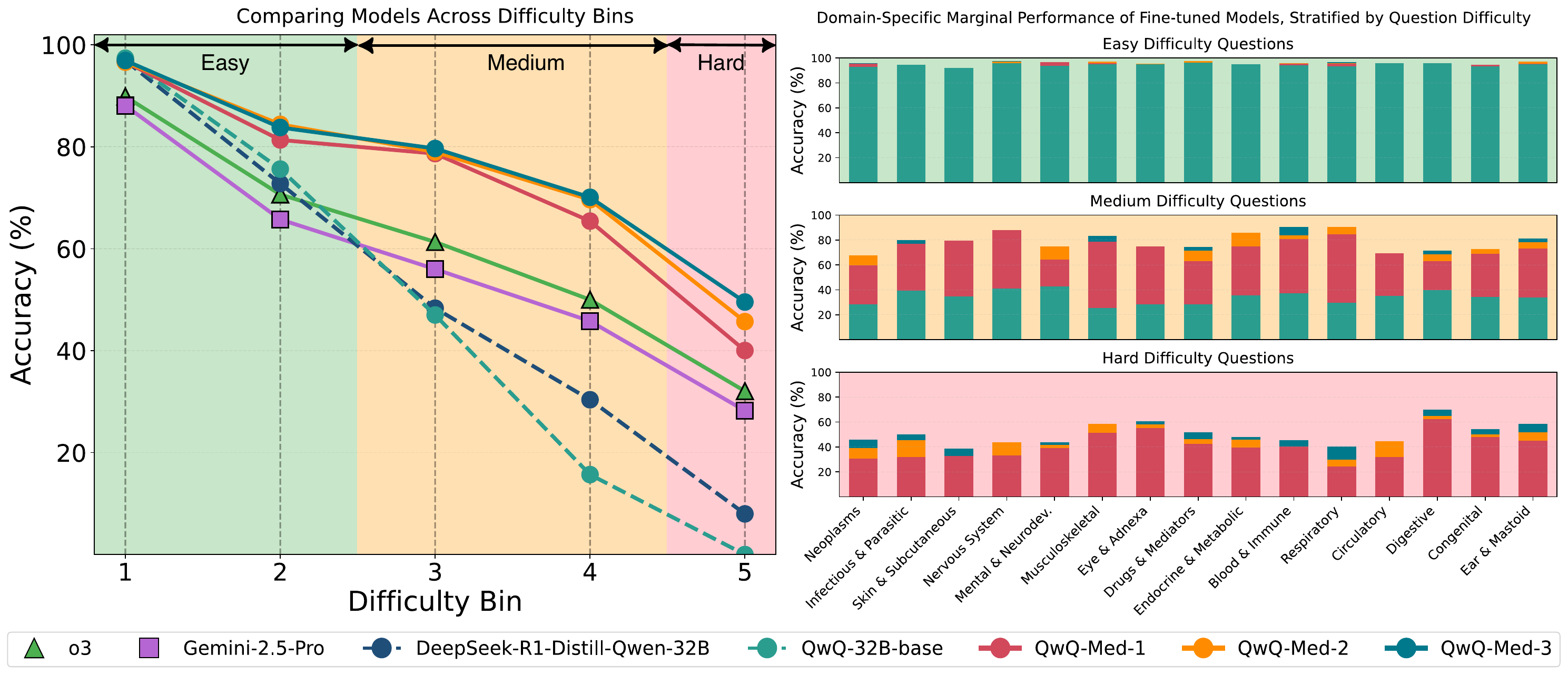}
    \captionsetup{belowskip=0pt,aboveskip=4pt}
    \caption{\textbf{Performance across task difficulty bins on ICD-Bench.} We construct a task-difficulty estimator using the pass$@1$ rate of the base model. Left: All models show declining accuracy with increasing difficulty, validating our pass@1-based difficulty estimator. Curriculum-tuned models exhibit improved robustness, with widening gains on increasingly challenging tasks. Right: Marginal gains across the curriculum tiers highlight that deep KG path exposure (QwQ-Med-3) is essential for solving the hardest questions, where the base model fails entirely.}
    \label{fig:difficulty}
\end{figure}

\textbf{(S3) Setup:} We stratify our ICD-Bench evaluation across difficulty bins and report in Fig.~\ref{fig:difficulty} the accuracy of curriculum-tuned models and previous baselines under the full inference budget setting. We find:

\textbf{(O3.1) Task difficulty is symmetrical across models.} All models exhibit a consistent drop in accuracy as task difficulty increases (Fig.~\ref{fig:difficulty}, left), validating our estimator. While open-source models like R1-Distilled Qwen and the base QwQ perform well on easy tasks, their performance significantly degrades on harder ones. Curriculum-tuned models, by contrast, outperform baselines across all bins, with the performance gap with the base widening at higher difficulties. This divergence suggests that our models are going beyond factual recall to internalizing structured reasoning capabilities that generalize to challenging scenarios while the base models exhibit surface-level knowledge but struggle to compose their reasoning to solve complex, unseen tasks.

\textbf{(O3.2) Curriculum-tuned models show greater marginal gains on harder tasks.} We further analyze the marginal contributions of successive curriculum-tuned models over the base (Fig.~\ref{fig:difficulty}, right). On easy tasks, most of the performance is already captured by the base model, indicating that KG primitives are inessential for most easy tasks. However, starting with medium-difficulty tasks, the base model contributes less than half the final performance, with QwQ-Med-1 bridging much of the remaining gap. Crucially, for the hardest tasks, the base model offers near-zero accuracy, with the final gains on these tasks coming almost entirely from QwQ-Med-3, indicating that exposure to deeper, compositional paths in the KG is necessary for bridging the gap between recall and reasoning, and for enabling models to tackle hard domain-specific reasoning tasks.

\begingroup
\definecolor{pam}{RGB}{215,48,39}
\definecolor{ph}{RGB}{255,127,0}
\definecolor{no}{RGB}{0,153,51}
\definecolor{relcause}{RGB}{117,107,177}
\definecolor{reltx}{RGB}{140,203,222}

\colorlet{pam30}{pam!30}
\colorlet{ph30}{ph!30}
\colorlet{no30}{no!30}
\colorlet{relcause30}{relcause!30}
\colorlet{reltx30}{reltx!30}

\newcommand{\hlpam}[1]{\sethlcolor{pam30}\hl{#1}}
\newcommand{\hlph}[1]{\sethlcolor{ph30}\hl{#1}}
\newcommand{\hlno}[1]{\sethlcolor{no30}\hl{#1}}
\newcommand{\hlrelcause}[1]{\sethlcolor{relcause30}\hl{#1}}
\newcommand{\hlreltx}[1]{\sethlcolor{reltx30}\hl{#1}}
\refstepcounter{boxcounter}
\begin{tcolorbox}[float,
    enhanced,            
    breakable=true,           
    colback=pink!20,
    colframe=black,
    title=Example~\theboxcounter: QwQ-Med-3 Response,
    width=\textwidth,
    label=qual_1,
    float, floatplacement=htb
]
\textbf{KG Path from Diseases of the Respiratory System:}\\[2pt]\hlpam{Pulmonary alveolar microlithiasis}
\(\xrightarrow{\text{\hlrelcause{maybe causes}}}\)
\hlph{Pulmonary hypertension}
\(\xrightarrow{\text{\hlreltx{treated by}}}\)
\hlno{Nitric oxide}\\[3pt]
\textbf{Question:}\\[3pt]
A 35-year-old male presents with progressive dyspnea, a loud P2, right ventricular heave, and sand-like pulmonary opacities. CT shows diffuse microliths. Echo reveals RV hypertrophy and pulmonary artery systolic pressure of 65 mmHg. Which of the following is the MOST appropriate therapy?\\[3pt]
\textbf{Options:}\\[3pt]
A. Inhaled corticosteroids \quad B. \textbf{Sildenafil} \quad C. N-acetylcysteine \quad D. Bronchodilators\\[3pt]
\textbf{QwQ-Med-3 Response:}\\[2pt]\texttt{<think>} Okay, let's break down this case step-by-step to figure out the best treatment. \textbf{1. Deconstruct the Presentation:} .... point clearly towards right heart strain, specifically \hlph{right ventricular failure due to pulmonary hypertension} .... \textbf{2. Interpret the imaging:} The chest X-ray shows those classic "sand-like" opacities. This is a huge clue! \hlpam{It almost always makes us think of Pulmonary Alveolar Microlithiasis (PAM)} .... \textbf{3. Connect the dots:} So, we have PAM confirmed. These little stones build up in the alveoli, basically filling them up .... \hlrelcause{Chronic lung disease like PAM is a well-known cause of Group 3 Pulmonary Hypertension, also called hypoxic pulmonary hypertension} .... Echocardiography shows right ventricular hypertrophy and an estimated pulmonary artery systolic pressure of 65 mmHg. .... \textbf{6. Evaluate the options:} .... B. Sildenafil: \hlno{This is a Phosphodiesterase-5 (PDE5) inhibitor}. \hlreltx{PDE5 inhibitors are specifically used to treat pulmonary hypertension by causing vasodilation (relaxing) of the pulmonary arteries.} .... standard therapy for PH, including Group 3 PH secondary to lung diseases like PAM.
\texttt{</think>} Final Answer: \textbf{B}

\end{tcolorbox}
\endgroup

\begingroup
\definecolor{gliosis}{RGB}{230,85,13}
\definecolor{cva}{RGB}{253,174,97}
\definecolor{ee}{RGB}{44,162,95}
\definecolor{relcause}{RGB}{117,107,177}
\definecolor{relrisk}{RGB}{140,203,222}
\colorlet{gliosis30}{gliosis!30}
\colorlet{cva30}{cva!30}
\colorlet{ee30}{ee!30}
\colorlet{relcause30}{relcause!30}
\colorlet{relrisk30}{relrisk!30}

\newcommand{\hlgliosis}[1]{\sethlcolor{gliosis30}\hl{#1}}
\newcommand{\hlcva}[1]{\sethlcolor{cva30}\hl{#1}}
\newcommand{\hlee}[1]{\sethlcolor{ee30}\hl{#1}}
\newcommand{\hlrelcause}[1]{\sethlcolor{relcause30}\hl{#1}}
\newcommand{\hlrelrisk}[1]{\sethlcolor{relrisk30}\hl{#1}}
\refstepcounter{boxcounter}
\begin{tcolorbox}[
    enhanced,            
    breakable=true,           
    colback=pink!20,
    colframe=black,
    title=Example~\theboxcounter: QwQ-Med-3 Response,
    width=\textwidth,
    label=qual_2
]
\textbf{KG Path from Diseases of the Nervous System:}\\[2pt]\hlee{Ethinylestradiol}$\xrightarrow{\text{\hlrelrisk{is a risk factor of}}}$\hlcva{Cerebrovascular accident}$\xrightarrow{\text{\hlrelcause{maybe causes}}}$ \hlgliosis{Gliosis}\\[3pt]
\textbf{Question:}\\[3pt]
 A 38-year-old woman with a history of migraines presents to her neurologist for follow-up. She has been taking an oral contraceptive containing \hlee{ethinylestradiol} for the past 10 years. ... MRI shows multiple small areas of T2 hyperintensity ... no evidence of hypertension or other cardiovascular risk factors. What is the MOST likely pathological process? \\[3pt]
\textbf{Options:}\\[3pt]
A. Demyelination \quad B. Apoptosis \quad C. \textbf{Reactive gliosis} \quad D. Amyloid angiopathy\\[3pt]
\textbf{QwQ-Med-3 Response:}\\[2pt]
\texttt{<think>} Okay, let's break down this question step-by-step .... \textbf{3. Consider the Role of Ethinylestradiol}. The patient has been on Ethinylestradiol for a long time. \hlee{Ethinylestradiol is an estrogen}. We need to think about the potential effects of long-term estrogen exposure .... \textbf{4. Connect Estrogen and Migraines:} \hlrelrisk{It's well-known that estrogen can have complex effects on migraines} .... \textbf{5. Link Migraines and White Matter Changes:} .... \hlrelrisk{recognized association between chronic migraines (especially migraine with aura) and the presence of these kinds of white matter lesions. The exact mechanism isn't fully understood} .... \textbf{6. Evaluate the Options based on the Pathophysiology:} ....  Reactive gliosis resulting from subclinical ischemic events: \hlcva{The white matter T2 hyperintensities represent areas of injury. This injury could be related to the migraine mechanism itself} .... \hlrelcause{The body's response to this injury involves glial cells proliferating and changing their properties} .... \hlgliosis{this is called gliosis.} .... \textbf{7. Conclude:} The most plausible explanation links the patient's chronic migraines (\hlrelrisk{potentially exacerbated by long-term ethinylestradiol use}) to \hlcva{subclinical vascular} or ischemic events in the brain. \hlrelcause{These events lead to tissue injury} and the subsequent \hlgliosis{reactive gliosis}, which manifests as T2 hyperintensities on MRI. \texttt{</think>} Final Answer: \textbf{C}

\end{tcolorbox}
\endgroup

\subsection{Disentangling the Effect of Curriculum Depth from Curriculum Diversity}
\label{sec:ablation}

The effectiveness of a curriculum-tuned model stems from training on reasoning data derived by exploring (1) multi-hop KG paths of different lengths over (2) diverse entities of the KG. We determine the contribution of each factor to the overall performance by incrementally ablating over the data mixture and size used for curriculum tuning.

\textbf{(S4) Setup:} We begin with an $8{,}000$-sample dataset derived solely from single-hop KG paths, representing a shallow but diverse baseline curriculum. To assess the effect of depth, we construct a second $8{,}000$-sample dataset consisting entirely of three-hop paths, holding diversity constant. A third variant samples uniformly from multi-hop paths of length ${1,2,3}$ using our proposed complexity-sampling procedure. Finally, we scale to $24{,}000$ samples via additional diversity sampling to maximize KG coverage. We fine-tune a model on each dataset under the same training FLOPs budget, and show their evaluation results on ICD-Bench in Fig.~\ref{fig:ablation}. Our major findings are:



\begin{figure}
    \centering
    \includegraphics[width=\linewidth]{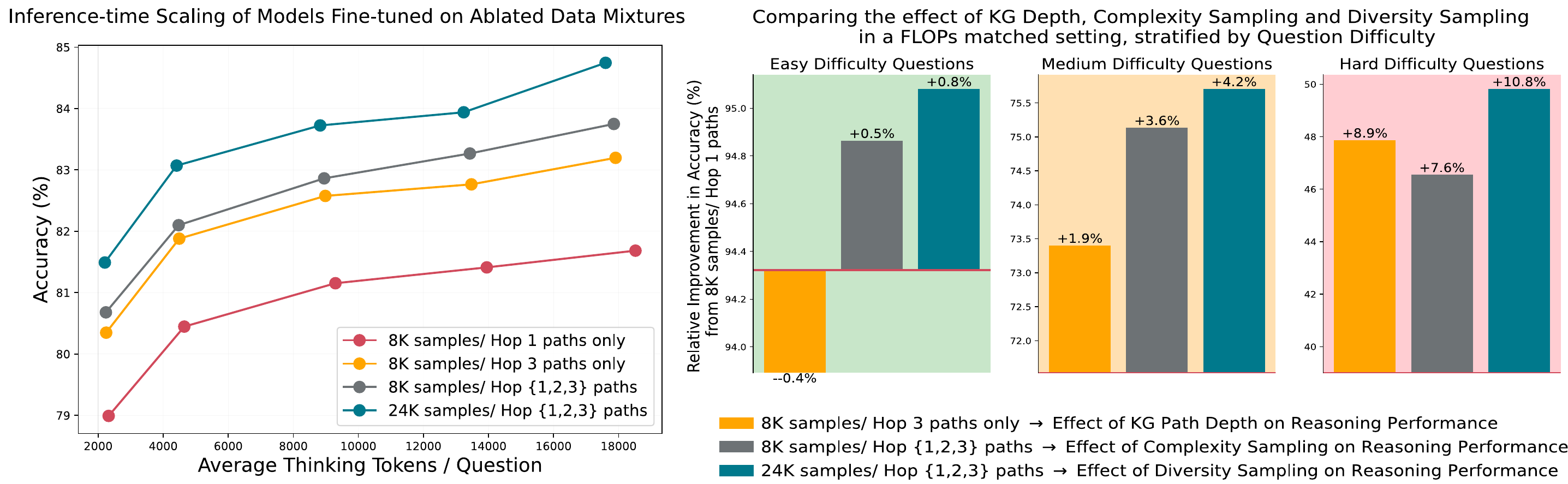}
    \captionsetup{belowskip=0pt,aboveskip=4pt}
    \caption{\textbf{Disentangling the effects of KG path depth, complexity sampling, and diversity sampling on curriculum-tuned model performance.} Left: Performance improves with increased thinking tokens as curricula incorporate deeper paths (3-hop), balanced path-length sampling, and greater diversity, demonstrating the additive benefits of each curation step. Right: Relative accuracy gains over a single-hop baseline stratified by task difficulty. Deeper paths are most helpful for challenging questions, balanced curricula are optimal for medium tasks, while easy tasks benefit most from diverse and balanced exposure.}
    \label{fig:ablation}
    \vspace{-1.5em}
\end{figure}

\textbf{(O4.1) Our full task-curation pipeline is important.} As shown in Fig.~\ref{fig:ablation} (left), each stage of our task-curation pipeline is crucial to overall performance. Incorporating deeper KG paths alone yields a substantial gain, while holding diversity constant, evident in the improvement from the single-hop to the three-hop dataset. However, depth alone is not sufficient: sampling a balanced mix of paths via our complexity sampling yields a further improvement, suggesting that exclusive exposure to long paths may lead to overfitting. Lastly, scaling the dataset to 24,000 samples through additional diversity sampling also yields improvement, though its impact is less pronounced than path depth.

\textbf{(O4.2) Compute-optimal KG depth depends on task difficulty.} In Fig.~\ref{fig:ablation} (right), we break down model performance on ICD-Bench by task difficulty, as estimated with the method presented in Section~\ref{sec:robustness}, comparing the relative gains of KG depth, complexity sampling, and diversity sampling over a single-hop baseline. Diversity sampling consistently improves performance across all difficulty levels. However, the optimal KG depth varies significantly with task difficulty. On easy tasks, using only three-hop chains slightly degrades performance, whereas the balanced dataset offers a modest improvement. For medium-difficulty tasks, a balanced mix of paths achieves near-optimal performance over maximizing path length. In contrast, on the hardest tasks, the three-hop-only dataset outperforms the balanced dataset. Together, these trends suggest that when task difficulty is known \textit{a priori}, the compute-optimal curricula depth should be composed accordingly: shallow paths suffice for easier tasks, moderate complexity benefits intermediate reasoning, and deep multi-hop traces become critical only at the hardest levels.

\subsection{Curriculum-Tuned Models Bridge the Recall-Reasoning Gap}
\label{sec:recall_reasoning}
Domain-specific reasoning hinges on being able to reliably recall relevant entities and relations, and then reason over them to reach a correct conclusion. We gain insight into the ability of our models to bridge recall and structured reasoning by diagnosing their generated thinking traces for alignment with the KG paths used to construct the question. This enables us to understand failure modes by disentangling errors due to inadequate recall from those arising due to erroneous inference. 

\begin{figure}
    \centering
    \includegraphics[width=\linewidth]{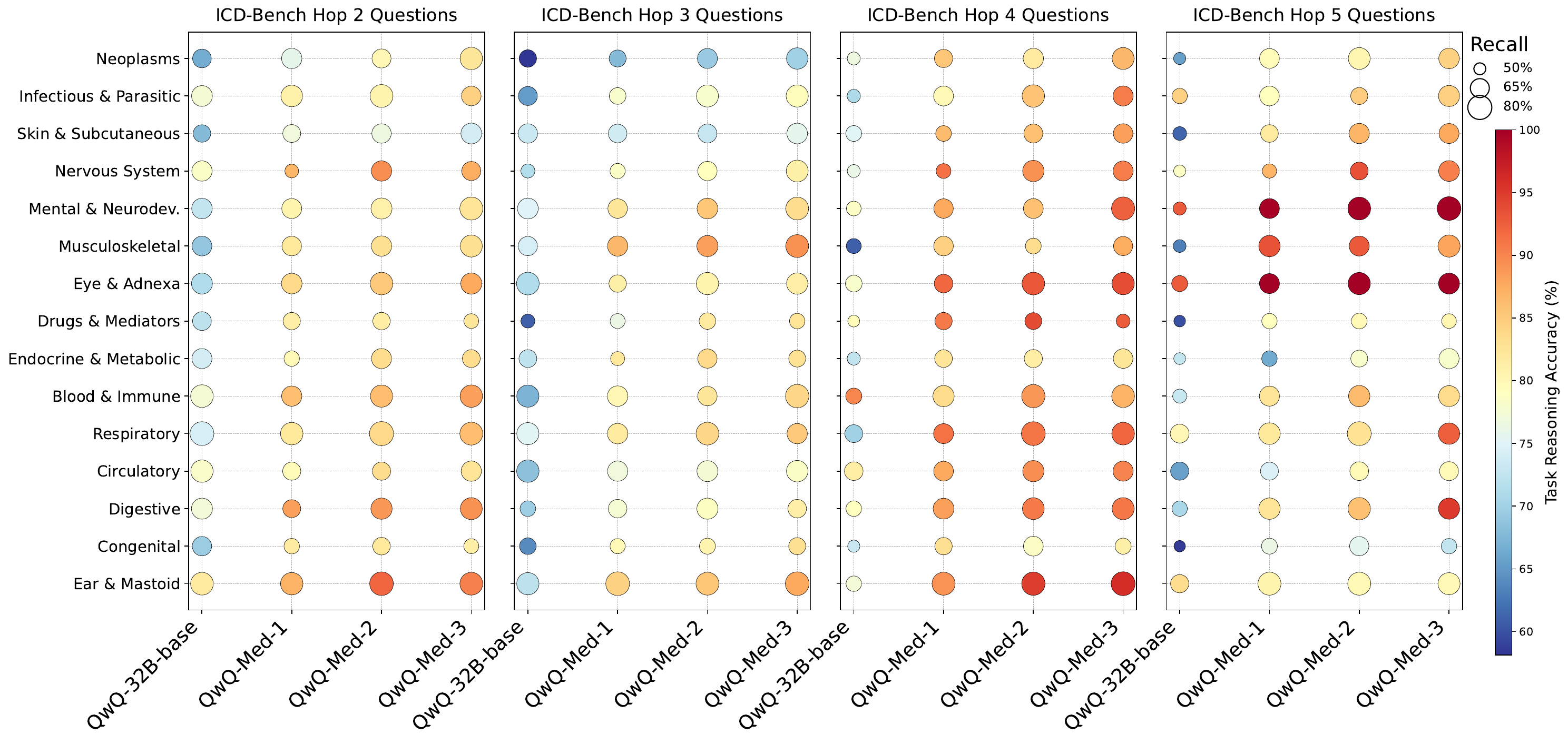}
    \captionsetup{belowskip=0pt,aboveskip=4pt}
    \caption{\textbf{Disentangling recall and reasoning performance across ICD-Bench tasks stratified by KG path length.} Dot size denotes recall (fraction of path entities recovered in the reasoning trace) and saturation reflects reasoning accuracy. Curriculum-tuned models demonstrate higher recall and effective reasoning across all hop levels, indicating successful use of KG primitives for reasoning. In contrast, the base model often retrieves relevant facts for shorter hop questions but fails to reason over them, revealing a surface-level understanding of the domain.}
    \label{fig:alignment}
\end{figure}

\textbf{(S5) Setup}: For each ICD-Bench task, we verbalize the ground-truth KG path into individual hop-level premises. An LLM judge independently evaluates whether each hop is explicitly delineated in the model's reasoning trace, enabling partial credit for alignment. We report recall as the fraction of hops utilized and reasoning efficacy as the performance accuracy. In Fig.~\ref{fig:alignment}, we stratify ICD-Bench by KG hop lengths and report both metrics across hop-wise subsets. We observe:

\textbf{(O5.1) Curriculum-tuned models effectively utilize recalled paths.} As shown in Fig.~\ref{fig:alignment}, curriculum-tuned models produce larger, more saturated dots across hop levels, reflecting both strong recall and effective reasoning. This indicates that these models are not merely retrieving KG primitives but effectively leveraging them to perform multi-step inference. Conversely, we also observe smaller, desaturated dots in certain categories (e.g., Drugs and Mediators), where even curriculum-tuned models struggle to reason correctly due to insufficient recall. These failure points underscore the importance of reliable recall for downstream reasoning, suggesting that model performance could further improve with more diverse training examples. 

\textbf{(O5.2) Base model can recall but fails to reason over retrieved knowledge.} In contrast, the base model shows a notable disconnect between recall and reasoning. On two-hop and even some three-hop tasks, it retrieves relevant KG hops at moderate rates, but its reasoning accuracy remains significantly lower. This highlights a failure mode that indicates the base model possesses surface knowledge relevant to the task, yet struggles to integrate it into a coherent reasoning trace. On questions requiring longer reasoning paths (e.g., four- and five-hop chains), both recall and reasoning degrade sharply, indicating that the base model lacks the structural inductive biases to generalize beyond shallow retrieval.

\subsection{Evaluation on Medical QA Benchmarks beyond the Original KG}
\label{sec:other_benchmarks}
To evaluate generalization beyond the scope of the KG, we assess our strongest model, QwQ-Med-3, on a suite of established medical QA benchmarks. Collectively, these datasets span a range of subdomains that provide a comprehensive evaluation for both robustness and generalization.

\textbf{(S6) Setup}: We benchmark the performance of QwQ-Med-3 against state-of-the-art open-source models: medical models like Meerkat \cite{Kim2025SmallTextbooks} and MedGemma \cite{Google2025MedGemmaDevelopers}, and general reasoning models like Deepseek-Distill-Qwen, Qwen 3\cite{yang2025qwen3technicalreport}, and Sky-T1 \cite{sky_t1_2025}. We compare them across four widely-used benchmarks: MedQA \cite{Jin2020WhatExams}, MedMCQA \cite{Pal2022MedMCQAAnswering}, MMLU-Med subset \cite{Hendrycks2020MeasuringUnderstanding}, and PubMedQA \cite{jin2019pubmedqa}. Each model is evaluated under identical precision settings and standard accuracy metrics (See Table \ref{tab:hyperparams} in Appendix~\ref{app:external}). The results are summarized in Table \ref{tab:medical_models_all}.

\textbf{(O6.1) Curriculum-tuned models reliably transfer acquired KG primitives.} Our curriculum-tuned model, QwQ-Med-3, demonstrates competitive or improved performance on external benchmarks, suggesting that the bottom-up primitives acquired through KG-grounded training generalize to tasks beyond the original curriculum. However, these benchmarks, derived from medical board-style questions, primarily assess factual recall rather than structured reasoning. While such recall is essential, our curriculum tuning enables models to go beyond isolated facts, extending this knowledge into structured reasoning by learning to compose across facts. In contrast, baseline models, despite performing reasonably well on recall-based tasks, struggle to extend their recall capability effectively to more compositional reasoning, revealing the limits of their surface-level understanding.

\begin{table}[!htb]
  \centering
  \setlength{\tabcolsep}{6pt}
  \renewcommand{\arraystretch}{1.3}

  \begin{tabular}{@{}l*{6}{c}@{}}
    \toprule
    Model & Size & \makecell{\# Training \\ Examples} &
    \makecell{MedQA\\USMLE} &
    \makecell{PubMed\\QA} &
    \makecell{Med\\MCQA} &
    \makecell{MMLU\\Med-Subset} \\

    \midrule
    \multicolumn{7}{c}{\textbf{Open-Source Models}} \\[2pt]
    \model{MedGemma}\footnotemark       & 27B  & --   & 60.49 & 70.40 & 63.42 & 78.46 \\
    \model{Meerkat}                     & 70B  & 441K & 78.95 & 77.40 & 68.42 & 83.74 \\
    \model{Deepseek-R1-Distill-Qwen}    & 32B  & 800K & 74.31 & 76.00 & 62.23 & 85.08 \\
    \model{Qwen3}                       & 32B  & --   & 64.34 & 65.40 & 55.63 & 69.53 \\
    \model{Sky-T1}                      & 32B  & 17K  & 70.78 & 67.40 & 63.52 & 84.91 \\
    \model{QwQ}                         & 32B  & --   & \underline{85.62} & 71.00 & 69.26 & 90.46 \\
    \model{QwQ/parallel-scaling}        & 32B  & --   & \textbf{87.09} & \underline{78.02}   & \underline{71.62} & \underline{91.56} \\

    \midrule
    \multicolumn{7}{c}{\textbf{Our Curriculum-Tuned Model}} \\[2pt]
    \model{QwQ-Med-3}                   & 32B  & 24K & 82.72 & 76.00 & 71.03 & 90.64 \\
    \model{QwQ-Med-3/parallel-scal.}    & 32B  & 24K & 85.39 & \textbf{78.19} & \textbf{73.25} &  \textbf{92.90}\\

    \bottomrule
  \end{tabular}
\vspace{1em}
\captionsetup{belowskip=0pt,aboveskip=4pt}
  \caption{\textbf{Benchmarking our curriculum-tuned model against open-source models.} The bottom four rows compare the base QwQ model with our curriculum-tuned variant, including results obtained under inference-time scaling of the base and fine-tuned model. Best performance on a benchmark is highlighted in  \textbf{bold}, with second-best performance \underline{underlined}. Expanded results are presented in~\autoref{tab:medical_models} in~\autoref{app:external}.}
  \label{tab:medical_models_all}
\end{table}

\section{Related Work}

\textbf{Reasoning with LLMs and KGs.} Despite excelling in many natural language tasks, LLMs often struggle with complex reasoning and lack in-depth knowledge, often hallucinating facts in critical domains \cite{Yu2025BenchmarkingModels, Longwell2024PerformanceQuestions}. To mitigate this, researchers have integrated KGs as structured external sources to improve an LLM's reasoning and factual recall \cite{Yasunaga2021QA-GNN:Answering, Su2024KGARevion:QA}. In the medical domain, UMLS-based KGs have improved clinical and diagnostic reasoning, with adapter-based approaches effectively injecting UMLS knowledge into biomedical QA models \cite{Gao2024LargePrediction, Bodenreider2004TheTerminology, Park2023SelectiveAnswering, Schmidgall2024AgentClinic:Environments}. Longitudinal studies have explored the use of smaller models, such as Graph Neural Networks or Long Short-Term Memory, to augment a larger model's reasoning \cite{Chen2020TowardNetworks, Zhang2022SubgraphAnswering, Liu2024ExploreGraph}. Although these methods have achieved incremental success, their performance remains constrained due to the reliance on smaller, less powerful models. Contrary to the existing techniques that use Graph Retrieval-Augmented Generation to build an LLM+KG pipeline and rely on static external retrievers to obtain current facts \cite{Liang2022AMultimodal}, we propose grounding the model in intricate relationships during the pre-/post-training phase itself. Integration of knowledge bottom-up paves the way to building of superintelligent models capable of complex multi-hop reasoning.

\textbf{Synthetic Data Curation for Instruction Tuning.} Aligning LLMs to complex tasks via instruction tuning is limited by the scarcity of high-quality, human-annotated instruction-response pairs. To overcome this, researchers curate synthetic datasets with powerful base models \cite{Zhang2023AlpaCare:Instruction-tunedApplication, Wang2022Self-Instruct:Instructions}.  Ref.~\cite{Chen2024HuatuoGPT-o1LLMs} illustrates this approach by introducing a GPT-4o-generated medical chain-of-thought (CoT) dataset with over 20K QA pairs; however, the dataset raises concerns about potential hallucinations in critical domains. Hybrid techniques, including a subset of expert-written examples, have achieved notable gains in domain-specific performance \cite{Nazar2025HowPerspective, Zhang2023AlpaCare:Instruction-tunedApplication}. To further ensure the quality of synthetic data samples, recent work leverages structured, verifiable sources such as textbooks, excerpts, to generate CoT reasoning chains for existing questions \cite{Kim2025SmallTextbooks} to boost the accuracy of small LMs. Ref.~\cite{Tatarinov2025KG-QAGen:Evaluation} uses a query-based method (SPARQL) to extract QA pairs from a manually-annotated KG at multiple complexity levels for long-context tasks. In addition to a QA pair, our method leverages KGs to extract explicit reasoning paths towards the correct answer, thereby enhancing the quality of the generated dataset and imposing an inherent curriculum based on path length. 

\footnotetext{Reported accuracy scores in the model card for MedGemma are different, using inference-time scaling, the details of which are undocumented.} 

\textbf{Curriculum Learning.} Inspired by human pedagogy, Curriculum Learning (CL) is a training paradigm where models are progressively exposed to training examples arranged from easy to hard during pretraining to facilitate more effective learning \cite{Bengio2009CurriculumLearning, Soviany2022CurriculumSurvey}. Early theoretical work demonstrated that difficulty-based ordering yields faster convergence and better performance on downstream tasks \cite{Wang2024CurriculumModels, Bengio2009CurriculumLearning}. Just as medical students must learn anatomy and physiology before they can diagnose and treat complex diseases, a neural network must learn simple one-hop relational triples before complex multi-hop reasoning. Recent studies apply CL to LLMs, showing that exposure to incrementally harder questions enhances reasoning and instruction following \cite{Ma2025Problem-SolvingReasoning, Luo2024LetsModels, Thawakar2025LlamaV-o1:LLMs}. Notably, even small LMs can exhibit emergent multi-step reasoning when trained on carefully constructed curricula \cite{Fu2025CanGPT-2, Nair2024CurriculumModels}. We leverage the KG to generate training questions of increasing difficulty, using the number of hops as a proxy for complexity and conduct extensive experiments using diverse training recipes and scaling test-time compute to examine the role of CL in enabling reasoning depth and generalization in LMs (see Section \ref{experiments}).

Additional Related Work is presented in Appendix \ref{sec:add_related_work}.

\section{Discussion}

\textbf{From Neural Abstractions to Data Abstractions.} Our experimental findings underscore the importance of directly integrating reasoning traces derived from domain-specific primitives into the training data, rather than relying on the LM representations to learn them from examples that implicitly utilize, but do not make explicit, the underlying structure. We demonstrate the efficacy of this principle in the context of medicine, where a reliably curated KG readily provides bottom-up abstractions for synthesizing reasoning traces. Several recent works have also curated high-quality data using domain-specific abstractions from formal languages \cite{lin2025goedelproverfrontiermodelopensource, chervonyi2025goldmedalist}, advanced examination questions \cite{muennighoff2025s1simpletesttimescaling}, and scientific forums \cite{yin2025scientificreasoningllmstraining}. As we saturate the usage of Internet text for training data \cite{villalobos2024rundatalimitsllm}, designing domain-specific data abstractions that can seamlessly interface with natural language to synthesize high-quality training tasks is a promising direction. 

\textbf{Training/Inference Energy Cost Reduction.}
LLMs incur exorbitant energy costs during both training and inference. However, since the LMs that are fine-tuned for superintelligence can potentially be much smaller, their fine-tuning and inference energy costs can also be substantially reduced.
Relying on a domain-specific architecture trained on abstracted data scaffolds to elicit high-quality reasoning, as opposed to a large architecture trained on unstructured Internet text, offers other efficiencies, such as requiring fewer inference tokens
to achieve superintelligent expertise.

\textbf{Bottom-up Primitives as Verifiable Rewards.}  Recent advances in reinforcement learning (RL) with LLMs have demonstrated success when guided by verifiable rewards \cite{deepseek2025r1,lambert2024tulu3,yue2025doesreinforcementlearningreally}, enabling significant strides in reasoning. While our current approach relies on SFT over full KG paths to instill structured reasoning, the same setup can be naturally reframed as an RL problem. In this view, each KG primitive along a path functions as a localized verifier, providing a dense reward signal whenever the model correctly recalls or traverses a valid relational edge. This transforms the KG into a fully simulatable training environment, where reasoning agents can be optimized not only for end-task correctness but also for intermediate trace fidelity. Such a paradigm opens promising avenues for training superintelligent systems in domains where high-quality, bottom-up abstractions enable precise reward shaping.

\textbf{Scalable Neurosymbolic Reasoning.} A persistent criticism of deep learning systems is their inability to perform robust symbolic reasoning, as they primarily perform high-dimensional statistical interpolation over training distributions rather than explicitly represent and manipulate structured knowledge, causal relationships, or compositional rules of a domain \cite{marcus2025generativeai}. Exact symbolic inference over large graphs is computationally intractable, and prior neurosymbolic systems, which rely on graph search or logic engines, have struggled to scale beyond toy domains. Our approach offers a scalable alternative by \textit{neuralizing} symbolic reasoning primitives. KG paths encode discrete structure, while curriculum tuning on tasks derived from these paths teaches an LM to internalize compositional reasoning over the graph without explicitly enumerating all paths. This results in a differentiable approximation to symbolic reasoning that remains grounded in verifiable primitives, enabling tractable multi-hop inference within the reasoning trace itself.

\textbf{Artificial General Intelligence (AGI) as Recursively Composable Bottom-Up Superintelligences.}
The dominant approach to AGI \cite{ai2027} today centers around scaling large monolithic architectures on domain-agnostic corpora to serve as a universal reasoning substrate across a broad spectrum of tasks.  Our work lays the foundation for an alternative perspective in which general intelligence is an emergent property of a modular system of interacting superintelligent agents \cite{minsky_societyofmind1986}. In this imagined system, each agent can (1) specialize in a domain by learning from domain-specific abstractions (e.g., KGs) and (2) learn to communicate or hand off subproblems to adjacent specialists, forming a collaborative mesh of expertise. At inference time, complex tasks can then be decomposed into subtasks aligned with these specialized agents, with their outputs recursively composed along the agent-level compute graph to produce a coherent solution. This compositional model of AGI will require engineering domain-specific verifiable primitives that are functionally \textit{local} to the agent as well as simulatable environments that allow \textit{global} interactions to emerge from \textit{local} primitives.

\textbf{Limitations.} While our work demonstrates promising results based on using a KG as a scaffold for deriving structured reasoning data, several constraints remain. First, the KG can be utilized beyond training to learn process reward models (PRM) \cite{lightman2023letsverifystepstep} from KG primitives and significantly improve inference-time scaling with PRM-guided search \cite{snell2024scalingllmtesttimecompute}.  Second, the underlying KG, despite providing a reliable structure over domain primitives, contains a closed vocabulary that constrains the conceptual coverage of the learned data abstractions. This limitation could be addressed by rigorously curating dense and high-quality KGs that cover diverse concepts. Third, we limit our focus to generating closed-ended multiple choice question tasks. A significant challenge lies in being able to generate open-ended tasks from a KG that can be reliably transferred to real-world use cases \cite{microsoft2025medicalsuperintelligence}.  Fourth, our difficulty heuristic utilizes oracle answers to estimate task difficulty. Reliably learning a model-based difficulty metric without ground truth answers can be useful. Finally, we demonstrate the efficacy of our method in medicine where a reliable KG is available and its generalizability to other domains (e.g., law, banking), that lack canonical KGs or standardized abstractions, remains to be fully validated. 

\section{Conclusion} 
We introduced a novel task-synthesis framework that traverses structured paths on a KG to generate reasoning tasks that directly abstract domain-specific primitives. We also introduced ICD-Bench, a new evaluation suite designed to quantify domain-specific reasoning abilities over diverse medical domains. Using our approach, we curated a bottom-up curriculum of $24,000$ medical reasoning tasks and fine-tuned QwQ-32B on our dataset, resulting in our curriculum-tuned QwQ-Med-3 model. This model outperforms other reasoning baselines across ICD-Bench and other established benchmarks. 

\begin{ack}


This work was supported by NSF under Grant No. CNS-2216746. The experiments reported
in this paper were performed on the computational resources managed and supported by Princeton Research Computing and the Princeton Language and Intelligence Initiative at Princeton University.
\end{ack}

\input{main.bbl}
\medskip


\appendix
\newpage
\section*{Appendices}

\section{Generating Grounded Question-Answering Tasks Using a KG}

\subsection{UMLS KG Details}
\begin{figure}[!hbt]
    \centering
    \includegraphics[width=\linewidth]{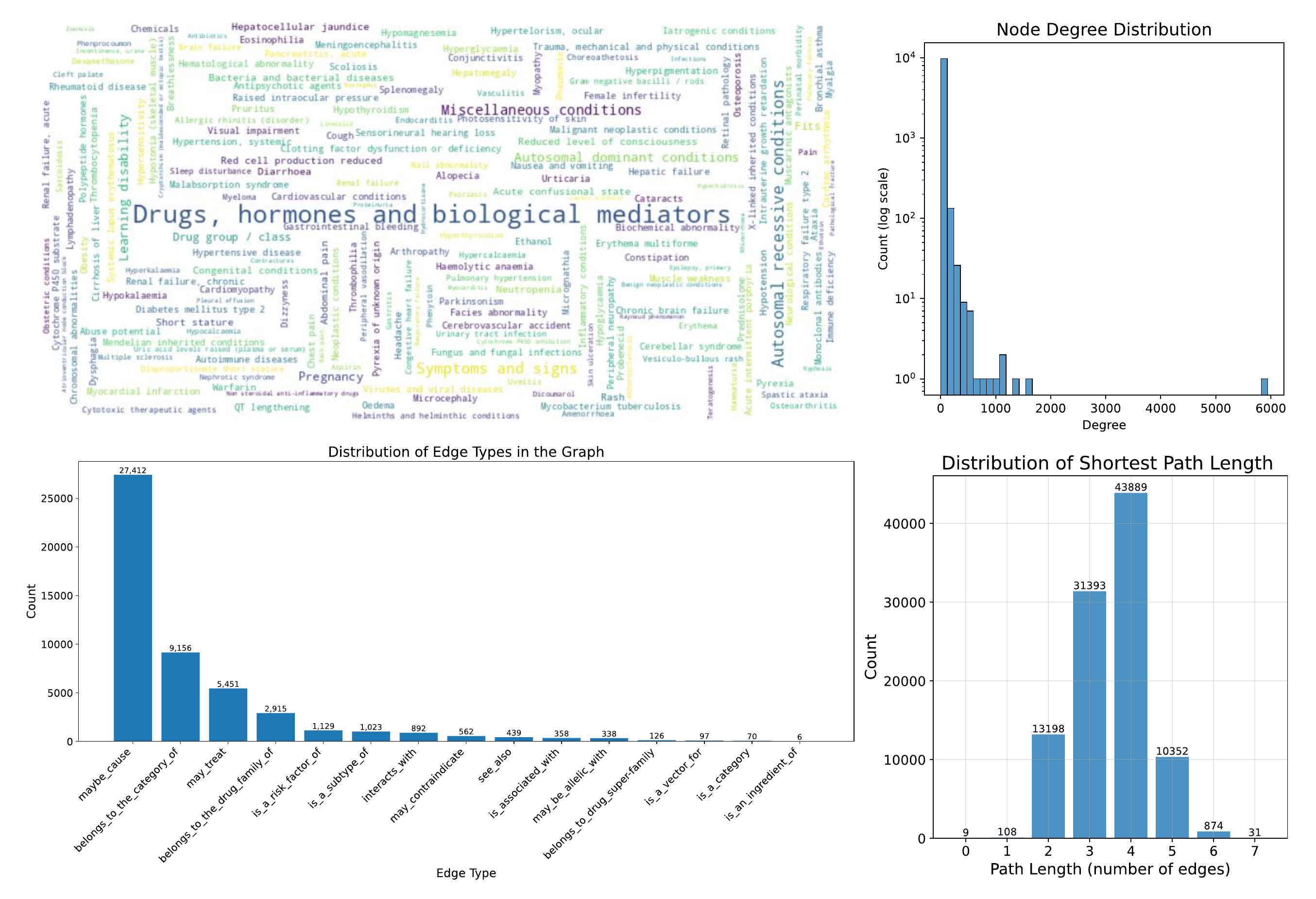}
    \caption{\textbf{Distributional statistics of the KG.} Top row: On the left, we visualize the top-$200$ nodes with the highest degree in a word cloud. On the right, we plot the histogram of node degrees. The KG is fairly sparse, and the weight is significantly concentrated on single-degree nodes. Bottom: On the left, we show the breakdown of the edges across their relation types. On the right, we randomly sample 100,000 node pairs and measure the shortest path distance between them. Most nodes are $\sim4$ hops away from each other, as a result of the sparsity of the KG.} 
    \label{fig:kg_stats}
\end{figure}

We use a KG constructed by past work \cite{Yasunaga2021QA-GNN:Answering} that integrates the Disease Database portion of the Unified Medical Language System (UMLS) \cite{Bodenreider2004TheTerminology} and DrugBank \cite{Wishart2018DrugBank}. The resultant KG contains $9{,}958$ nodes and $44{,}561$ edges. We utilized all edge relation types of the KG, with the exception of ``belongs to the category of,'' ``is a category,'' and ``is a subtype of,'' to avoid generating tasks that test UMLS taxonomy-based recall.  Fig.~\ref{fig:kg_stats} shows the distributional statistics of the KG. 

\subsection{QA Generation Prompt}
\label{app:qa_generation_template}
Prompt~\ref{prompt_qa} details the template used to transform KG paths into QA pairs. Given a KG path, the template generates a reasoning task that links the initial entity to the final entity by traversing the intermediate relations.

\vspace{1cm}
\refstepcounter{promptcounter}
\begin{tcolorbox}[
    enhanced,            
    breakable=true,           
    colback=gray!20,
    colframe=black,
    title=Prompt~\thepromptcounter: QA Generation from KG path Prompt Template,
    width=\textwidth,
    label=prompt_qa]

Create a medical examination question for advanced medical students that tests the relationship between 
\{\#insert source entity\} and \{\# insert target entity\}.  The relationship is: \{\# insert KG path here\}. The question should:

\hspace{2em} 1. Be in multiple choice format (4 options)

\hspace{2em} 2. Require clinical reasoning along the relationship

\hspace{2em} 3. Include a brief clinical vignette

\hspace{2em} 4. Not directly mention the relationship in the question stem

\hspace{2em} 5. Have one clearly correct answer

Format:

\hspace{1em}<Question>

\hspace{3em}[Clinical Vignette]

\hspace{1em}</Question>

\hspace{1em}<Options>

\hspace{3em}A. [Option]

\hspace{3em}B. [Option]

\hspace{3em}C. [Option]

\hspace{3em}D. [Option]

\hspace{1em}</Options>

\hspace{1em}<Answer>: 

\hspace{3em}[Correct Option Letter] 

\hspace{1em}</Answer>
\end{tcolorbox}
\subsection{Task Generation Pipeline Details}
\label{app:task_generation_pipleline}
Prompt~\ref{prompt_template_trace} presents the template used to generate a thinking trace for a QA pair, conditioned on its corresponding KG path. Prompt~\ref{prompt_template_correctness} defines the template employed by the LLM grader to assess correctness. The grader evaluates the alignment between the answer, thinking trace, and KG context, using the full QA pair as input. The overall curriculum curation pseudo-code is presented in Algorithm~\ref{alg:curriculum}.

\vspace{1cm}
\refstepcounter{promptcounter}
\begin{tcolorbox}[
    enhanced,            
    breakable=true,           
    colback=gray!20,
    colframe=black,
    title=Prompt~\thepromptcounter: Thinking Trace Generation Prompt Template,
    width=\textwidth,
    label=prompt_template_trace
]

Generate a detailed explanation for the question: \{\#insert question and options\} 

Use the following context \{\#insert KG path here\}. The explanation should be:

\hspace{1em} 1. Detailed and include all the steps leading to the answer.

\hspace{1em} 2. You are to use the provided context to explain the relationship between the concepts.

\hspace{1em} 3. Strictly do not mention that you are using a given context to generate the explanation.
\end{tcolorbox}
\refstepcounter{promptcounter}
\begin{tcolorbox}[
    enhanced,            
    breakable=true,           
    colback=gray!20,
    colframe=black,
    title=Prompt~\thepromptcounter: Correctness Filtering Prompt Template,
    width=\textwidth,
    label=prompt_template_correctness
]

You are a medical examiner. You are given a medical question along with an explanation and the answer. You have also been given a source context.

\hspace{1em}1. Judge whether the question and answer are logically correct and medically accurate, and follow the source. If there is an explanation, also evaluate whether the explanation follows from the source to reach the correct answer.
       
\hspace{1em}2. Respond with only "Yes" or "No".
        
Format your response exactly like this: ``Correct: [Yes/No]''

Question: \{\# insert question and options here\}

Explanation:  \{\# insert thinking trace here\}

Answer: \{\# insert answer option letter here\}

Source Context: \{\# insert KG path here\}
\end{tcolorbox}

\clearpage
\begin{algorithm}[H]
\caption{Curriculum Curation Pseudo-code}
\label{alg:curriculum}
\KwIn{Knowledge Graph $\mathcal{G}$, Max Path Length $N$, QA template $\mathcal{T}_{\text{QA}}$, Thinking Model Prompt $\mathcal{T}_{\text{think}}$, Correctness Filtering Prompt $\mathcal{T}_{\text{verify}}$, TotalSamples}
\KwOut{High-quality QA pairs with grounded thinking traces}

Initialize node frequency table $\{f_i = 0\}$ for each $i \in \mathcal{G}$\;
Initialize dataset $\mathcal{D} \leftarrow \emptyset$\;

\While{size($\mathcal{D}$) != TotalSamples}{
    Sample source node $h_0 \sim \text{InverseFreqSampling}(f_i)$ \textbf{// Diversity Sampling}\;

    Sample path length $L \sim \text{Uniform}(\{1, \dots, N\})$ \textbf{// Complexity Sampling}\;

    Sample path $p^L = \left(h_0, r_1, h_1, \dots, r_L, h_L\right)$ from $\mathcal{G}$  \textbf{// KG Path Sampling}\;
    
    Generate $(q, A, \text{Options}) = \texttt{Gemini-2.0-Flash}(\mathcal{T}_{\text{QA}}(h_0, h_L, p^L))$ \textbf{// QA Generation}\;

    \textbf{// Quality Filtering}\;
    \If{Invalid formatting, distractors, or missing fields}{
        \textbf{continue}\;
    }

    Generate trace $T = \texttt{Gemini-2.5-Pro}(\mathcal{T}_{\text{think}}(q, \text{Options}, p^L))$ \textbf{// Thinking Trace Generation}\;

    \textbf{// Correctness Filtering with Dual LLMs}\;
    verdict$_1$ = \texttt{Gemini-2.0-Flash}($\mathcal{T}_{\text{verify}}(q, A, T, p^L)$)\;
    verdict$_2$ = \texttt{Qwen-2.5-72B}($\mathcal{T}_{\text{verify}}(q, A, T, p^L)$)\;

    \If{verdict$_1$ == pass \textbf{and} verdict$_2$ == pass}{
        Add $(q, A, \text{Options}, T)$ to $\mathcal{D}$\;
        Increment frequency count $f_i$ for all nodes in $p^L$\;
    }
}
\Return $\mathcal{D}$
\end{algorithm}

\section{UMLS-Curated Curriculum Details}
\label{app:UMLS_curriculum}
Examples~\ref{eg_umls_1} and ~\ref{eg_umls_2} illustrate generated samples of the thinking trace in our UMLS-curated curriculum, used to fine-tune our curriculum-tuned models. The traces are generated by \texttt{Gemini 2.5 pro} using Prompt~\ref{prompt_template_trace}. We summarize our dataset in Table~\ref {tab:umls_curriculum_stats} by aligning questions to the ICD taxonomy. \\[2pt] 

\refstepcounter{boxcounter}
\begin{tcolorbox}[
    enhanced,            
    breakable=true,           
    colback=pink!20,
    colframe=black,
    title=Example~\theboxcounter: Thinking trace from the UMLS-curated training curriculum,
    width=\textwidth,
    label=eg_umls_1
]

\textbf{KG-path}:

Au-Kline syndrome $\xrightarrow{\text{maybe cause}}$ Blepharoptosis $\xleftarrow{\text{maybe cause}}$ Turner syndrome $\xrightarrow{\text{maybe cause}}$ Cubitus valgus\\[2pt]
\textbf{Question:}\\[1pt]
A 16-year-old female presents to your clinic with a long-standing history of bilateral ptosis. On examination, you note mild developmental delay, subtle dysmorphic facial features, and a history of recurrent upper respiratory infections. She denies any muscle weakness or fatigability. Her height is in the 25th percentile for her age. You order genetic testing, and the results are pending. Based on the most likely underlying diagnosis, which of the following physical examination findings would you be MOST likely to identify?\\[2pt]
\textbf{Options:}\\[3pt]
A. Proximal muscle weakness and elevated creatine kinase \quad B. Webbed neck and short stature\quad C. Down-sloping palpebral fissures and epicanthal folds \quad \textbf{D. Increased carrying angle of the elbows}\\[1pt]

\textbf{Gemini 2.5 Pro Thinking Trace:}\\[1pt]

Okay, let's break this down like we're figuring it out together for rounds.
\begin{enumerate}[leftmargin=1.2em]

\item \textbf{Analyze the Patient's Presentation:} We have a 16-year-old female with a few key things going on:
\begin{itemize}
    \item \textbf{Bilateral ptosis (droopy eyelids):} This has been there a long time.
    \item \textbf{Mild developmental delay \& subtle dysmorphic features:} Suggests a possible underlying genetic syndrome.
    \item \textbf{Recurrent URIs:} Could be related to immune function or anatomical issues (like Eustachian tube problems).
    \item \textbf{No muscle weakness/fatigability:} This makes things like myasthenia gravis or progressive muscular dystrophies less likely as the \emph{primary} issue explaining the whole picture.
    \item \textbf{Height at 25th percentile:} She's on the shorter side for her age, but not profoundly short. Still, short stature in the context of other features is significant.
\end{itemize}

\item \textbf{Consider the Differential Diagnosis based on the key features:} The combination of ptosis, short stature (relative), developmental delay, and dysmorphic features screams "genetic syndrome." Let's look at the options through this lens:
\begin{itemize}
    \item \textbf{A. Proximal muscle weakness and elevated CK:} This points towards a primary muscle disease (myopathy/dystrophy). While some syndromes with ptosis \emph{can} have muscle involvement (like mitochondrial myopathies or myotonic dystrophy), the patient specifically \emph{denies} weakness or fatigability, making this unlikely to be the \emph{most} likely additional finding.
    
    \item \textbf{C. Down-sloping palpebral fissures and epicanthal folds:} These are very characteristic features of Down syndrome (Trisomy 21). Down syndrome does involve developmental delay, dysmorphic features, and increased risk of infections. However, ptosis isn't typically listed as a \emph{primary} or defining feature, although it can occur. Also, the question presents \emph{ptosis} as a major complaint alongside the other features, suggesting it's central to the suspected diagnosis.
    
    \item \textbf{B. Webbed neck and short stature \& D. Increased carrying angle of the elbows (Cubitus valgus):} Both of these are strongly associated with \textbf{Turner Syndrome (TS)}. Let's see if TS fits the overall picture:
    \begin{itemize}
        \item \textbf{Cause:} Usually 45,X karyotype or mosaicism. Affects females.
        \item \textbf{Ptosis:} Yes, ptosis can occur in Turner Syndrome.
        \item \textbf{Short Stature:} This is a hallmark feature. While often below the 5th percentile, presentation can vary, and 25th percentile is certainly relatively short and fits.
        \item \textbf{Developmental Delay:} Usually, intelligence is normal in TS, but specific non-verbal learning disabilities are common and might be described as "mild developmental delay" in some contexts.
        \item \textbf{Dysmorphic Features:} TS has characteristic features like low posterior hairline, high-arched palate, micrognathia, and sometimes epicanthal folds (though maybe not typically down-sloping fissures). A webbed neck (pterygium colli) is classic (Option B). Skeletal anomalies like a broad chest and cubitus valgus (increased carrying angle at the elbow, Option D) are also very common.
        \item \textbf{Recurrent Infections:} Recurrent otitis media is common in TS due to Eustachian tube dysfunction, which might be reported by the family as recurrent URIs.
    \end{itemize}
\end{itemize}

\item \textbf{Choose the MOST Likely Finding:} So, Turner Syndrome seems like a very strong candidate diagnosis given the patient's features (female, ptosis, relative short stature, possible dysmorphism/DD). Both webbed neck (B) and increased carrying angle (D) are features of TS.
\begin{itemize}
    \item A webbed neck is a classic sign but isn't present in all individuals with TS (maybe 25--50\%) and can be subtle.
    \item An increased carrying angle (cubitus valgus) is a \emph{very} common skeletal finding in TS, reported in over 50\% (some sources say >70\%) of individuals. It's often readily apparent on physical exam when checking the arms.
    \item Therefore, between B and D, the increased carrying angle (cubitus valgus) is statistically more frequent and thus the \emph{most likely} additional finding you'd identify on examination if the underlying diagnosis is indeed Turner Syndrome.
\end{itemize}

\item \textbf{Final Conclusion:} Based on the patient's constellation of symptoms pointing strongly towards Turner Syndrome as the most likely underlying diagnosis, an increased carrying angle of the elbows (cubitus valgus) is the most probable associated physical finding among the choices provided.

\end{enumerate}

\end{tcolorbox}

\refstepcounter{boxcounter}
\begin{tcolorbox}[
    enhanced,            
    breakable,           
    colback=pink!20,
    colframe=black,
    title=Example~\theboxcounter: Thinking trace from the UMLS-curated training curriculum,
    width=\textwidth,
    label=eg_umls_2
]

\textbf{KG-path}:
Aripiprazole$\xrightarrow{\text{may treat}}$ Schizophrenia\\[2pt]
\textbf{Question:}\\[1pt]
A 24-year-old male with a history of psychosis presents to the emergency department with increasing agitation, disorganized speech, and paranoid ideation. He reports auditory hallucinations that command him to harm himself. His family confirms a previous diagnosis of a chronic mental illness and that he stopped taking his prescribed medication approximately 3 months ago due to feeling "like a zombie." Physical examination is unremarkable. After initial stabilization with a short-acting benzodiazepine, which of the following long-term medication strategies is MOST appropriate to address both the positive and negative symptoms while minimizing the risk of extrapyramidal side effects and significant weight gain?\\[2pt] 
\textbf{Options:}\\[3pt]
A. Haloperidol decanoate injection every 4 weeks. \quad B. Clozapine with weekly white blood cell count monitoring.
\quad C. Risperidone long-acting injection every 2 weeks.
\quad \textbf{D. Aripiprazole oral daily.}
\\[2pt]
\textbf{Gemini 2.5 Pro Thinking Trace:}\\[1pt]
Okay, let's break down this question step-by-step to figure out the best long-term medication strategy for this patient.
\begin{enumerate}[leftmargin=1.2em]
\item \textbf{Understand the Patient and Situation:} We have a young adult male (24 years old) experiencing an acute psychotic episode (agitation, disorganized speech, paranoia, command hallucinations). This is likely an exacerbation of a known chronic mental illness, probably schizophrenia spectrum disorder given the symptoms and age. A key factor is that he \emph{stopped} his previous medication because of side effects, describing feeling "like a zombie," which often implies sedation, cognitive slowing, or extrapyramidal symptoms (EPS) like Parkinsonism or akinesia. He's been stabilized short-term with a benzodiazepine, but now needs a long-term plan.

\item \textbf{Identify the Goals of Treatment:} The question specifically asks for a medication that:
\begin{itemize}
    \item Treats both positive symptoms (hallucinations, paranoia, disorganized thought) AND negative symptoms (which are common in chronic psychosis, like apathy, flat affect, alogia - though not explicitly described here, they are a target for treatment).
    \item Minimizes the risk of EPS (like Parkinsonism, dystonia, akathisia, tardive dyskinesia). This is crucial given his previous experience ("zombie" feeling).
    \item Minimizes the risk of significant weight gain (a common and problematic side effect of many antipsychotics).
\end{itemize}

\item \textbf{Evaluate the Options Based on Goals:}

\begin{itemize}

    \item \textbf{A. Haloperidol decanoate injection:} This is a long-acting injectable (LAI) form of a first-generation antipsychotic (FGA).
    \begin{itemize}
        \item \emph{Pros:} LAI helps with adherence, which is good since he stopped meds before. Effective for positive symptoms.
        \item \emph{Cons:} Haloperidol is notorious for causing EPS. It has limited effectiveness against negative symptoms. The "zombie" feeling could very well have been EPS from a previous similar drug, making this a poor choice for minimizing that specific risk.
        \item \emph{Conclusion:} Doesn't meet the criteria for minimizing EPS or treating negative symptoms well.
    \end{itemize}

    \item \textbf{B. Clozapine:} This is a second-generation antipsychotic (SGA), often considered the most effective.
    \begin{itemize}
        \item \emph{Pros:} Highly effective for both positive and negative symptoms, especially in treatment-resistant cases. Very low risk of EPS.
        \item \emph{Cons:} Requires rigorous weekly (initially) blood monitoring due to the risk of agranulocytosis (dangerously low white blood cells). It also carries a high risk of significant weight gain, sedation, constipation, and other side effects. It's typically reserved for patients who haven't responded adequately to at least two other antipsychotics.
        \item \emph{Conclusion:} While it has low EPS risk and treats both symptom types, the high risk of weight gain and the significant monitoring burden/safety concerns make it inappropriate as a first-line or second-line choice here. This patient hasn't necessarily failed other treatments yet.
    \end{itemize}

    \item \textbf{C. Risperidone long-acting injection:} This is an LAI form of an SGA.
    \begin{itemize}
        \item \emph{Pros:} LAI helps with adherence. Effective for positive symptoms and some effect on negative symptoms.
        \item \emph{Cons:} Among SGAs, risperidone has a relatively higher risk of causing EPS (especially at higher doses) and hyperprolactinemia (which can cause sexual dysfunction, gynecomastia). It's also associated with moderate-to-significant weight gain.
        \item \emph{Conclusion:} While better than haloperidol, it still carries a considerable risk for EPS and weight gain, which we want to minimize.
    \end{itemize}

    \item \textbf{D. Aripiprazole oral daily:} This is an SGA with a unique mechanism (D2 partial agonism).
    \begin{itemize}
        \item \emph{Pros:} Effective for positive symptoms and potentially beneficial for negative and cognitive symptoms. It has a \emph{lower} risk of EPS compared to FGAs and many other SGAs like risperidone (though akathisia can occur). Crucially, it has a \emph{lower} risk of significant weight gain and metabolic side effects compared to agents like clozapine, olanzapine, and risperidone. Aripiprazole is known to be generally less sedating for many patients.
        \item \emph{Cons:} It's an oral medication, so adherence could still be an issue. However, starting with an oral agent that has a better side-effect profile is a reasonable first step, addressing the \emph{reason} he stopped his previous medication. If adherence fails despite better tolerability, an LAI version of aripiprazole exists.
        \item \emph{Conclusion:} This option provides the best balance. It targets positive and negative symptoms, has a lower risk profile for the specific side effects we're concerned about (EPS and weight gain), and directly addresses the likely reason for his previous non-adherence (poor tolerability). Aripiprazole is often considered a good choice when trying to balance efficacy with minimizing metabolic side effects and EPS.
    \end{itemize}

\end{itemize}

\item \textbf{Final Decision:} Comparing the options, Aripiprazole (D) best fits the requirements laid out in the question: efficacy for positive/negative symptoms, minimized EPS risk, and minimized weight gain risk, making it the most appropriate long-term strategy for this patient given his history and presentation.

\end{enumerate}

\end{tcolorbox}

\begin{table}[htbp]
\hspace*{-1.5cm}
\centering
\small
\setlength{\tabcolsep}{2pt}
\renewcommand{\arraystretch}{1.2}
\begin{tabular}{@{}p{4.2cm}ccc p{6cm}@{}}
\toprule
\textbf{Category} & \textbf{Total Questions} & \textbf{Avg Thinking Tokens} & \textbf{Total Tokens} & \textbf{Most Frequent Entities (Top 5)} \\
\midrule
Certain infectious and parasitic diseases & 2588 & 1128.03 & 2919330 & 
\makecell[l]{Gram negative bacilli / rods,\\ Meningoencephalitis,
Labyrinthitis, \\ Mycobacterium tuberculosis, \\
Viral haemorrhagic fever} \\
\midrule
Neoplasms & 1862 & 1076.20 & 2003880 & 
\makecell[l]{Bronchogenic carcinoma, Melanoma, \\Hemangioma,
Myeloma, \\
Colorectal cancer} \\
\midrule
Blood/immunity disorders & 2425 & 1173.51 & 2845761 & 
\makecell[l]{Thrombocytopenia, Neutropenia,\\ Red cell production reduced,\\
Thrombophilia, Haemolytic anaemia} \\
\midrule
Endocrine/nutritional/metabolic & 3387 & 1198.75 & 4060163 &
\makecell[l]{Hyperglycaemia, Hypoglycaemia, Enzymes,\\
Obesity, Malabsorption syndrome} \\
\midrule
Mental and neurodevelopmental disorders & 1824 & 1069.55 & 1950856 &
\makecell[l]{Learning disability, Sleep disturbance, \\Labyrinthitis,
Involuntary muscular movements,\\ Deliberate self harm} \\
\midrule
Nervous system diseases & 4083 & 1118.38 & 4566330 &
\makecell[l]{Learning disability, Headache,\\ Acute confusional state,
Chronic brain failure, \\Cerebellar syndrome} \\
\midrule
Eye and adnexa diseases & 1974 & 1091.62 & 2154860 &
\makecell[l]{Retinal pathology, Cataracts, Corneal opacity,\\
Conjunctivitis, Eye pain} \\
\midrule
Ear and mastoid diseases & 1064 & 1035.77 & 1102058 &
\makecell[l]{Sensorineural hearing loss,\\ Conductive hearing loss,
Tinnitus,\\ Suppurative otitis media, Hearing loss} \\
\midrule
Circulatory system diseases & 3213 & 1130.10 & 3631027 &
\makecell[l]{Respiratory failure type 2, QT lengthening,\\
Pulmonary hypertension,\\ Spastic ataxia, Cardiomyopathy} \\
\midrule
Respiratory system diseases & 1663 & 1092.93 & 1817539 &
\makecell[l]{Respiratory failure type 2, Bronchial asthma,\\
Breathlessness, Pneumonia, Cough} \\
\midrule
Digestive system diseases & 2932 & 1108.66 & 3250596 &
\makecell[l]{Diarrhoea, Gastrointestinal bleeding,\\
Renal failure (chronic), Pyrexia of unknown origin,\\ Dysphagia} \\
\midrule
Skin and subcutaneous tissue diseases & 6646 & 1129.23 & 7504844 &
\makecell[l]{Hepatocellular jaundice, Headache, Pruritus,\\
Neutropenia, Hepatomegaly} \\
\midrule
Musculoskeletal/connective tissue diseases & 4218 & 1128.49 & 4759982 &
\makecell[l]{Fits, Arthropathy, Muscle weakness,\\
Rheumatoid disease, Myalgia} \\
\midrule
Congenital/chromosomal abnormalities & 4065 & 1145.85 & 4657897 &
\makecell[l]{Microcephaly, Micrognathia,\\
Deafness onychodystrophy syndrome, Cleft palate, \\ Syndactyly} \\
\midrule
Drugs, hormones, and mediators & 2163 & 1138.08 & 2461671 &
\makecell[l]{Cytochrome P450 substrate, \\Cytotoxic therapeutic agents,
Dexamethasone,\\ Cyclophosphamide, Dicoumarol} \\
\bottomrule
\end{tabular}
\vspace{0.2em}
\caption{ \textbf{Summary of our generated curriculum:} We categorize each generated question into one or more ICD-10 categories by checking whether its KG path includes an entity from that category. For each category, we compute the total token count and the average length in the thinking traces, using the QwQ tokenizer. In addition, we identify the most frequent entities in each category to serve as representative keywords. }
\label{tab:umls_curriculum_stats}
\end{table}

\clearpage

\section{Supervised Fine-Tuning Setup}
\label{app:training_details}
\begin{figure}[!hbt]
    \centering
    \includegraphics[width=0.75\linewidth]{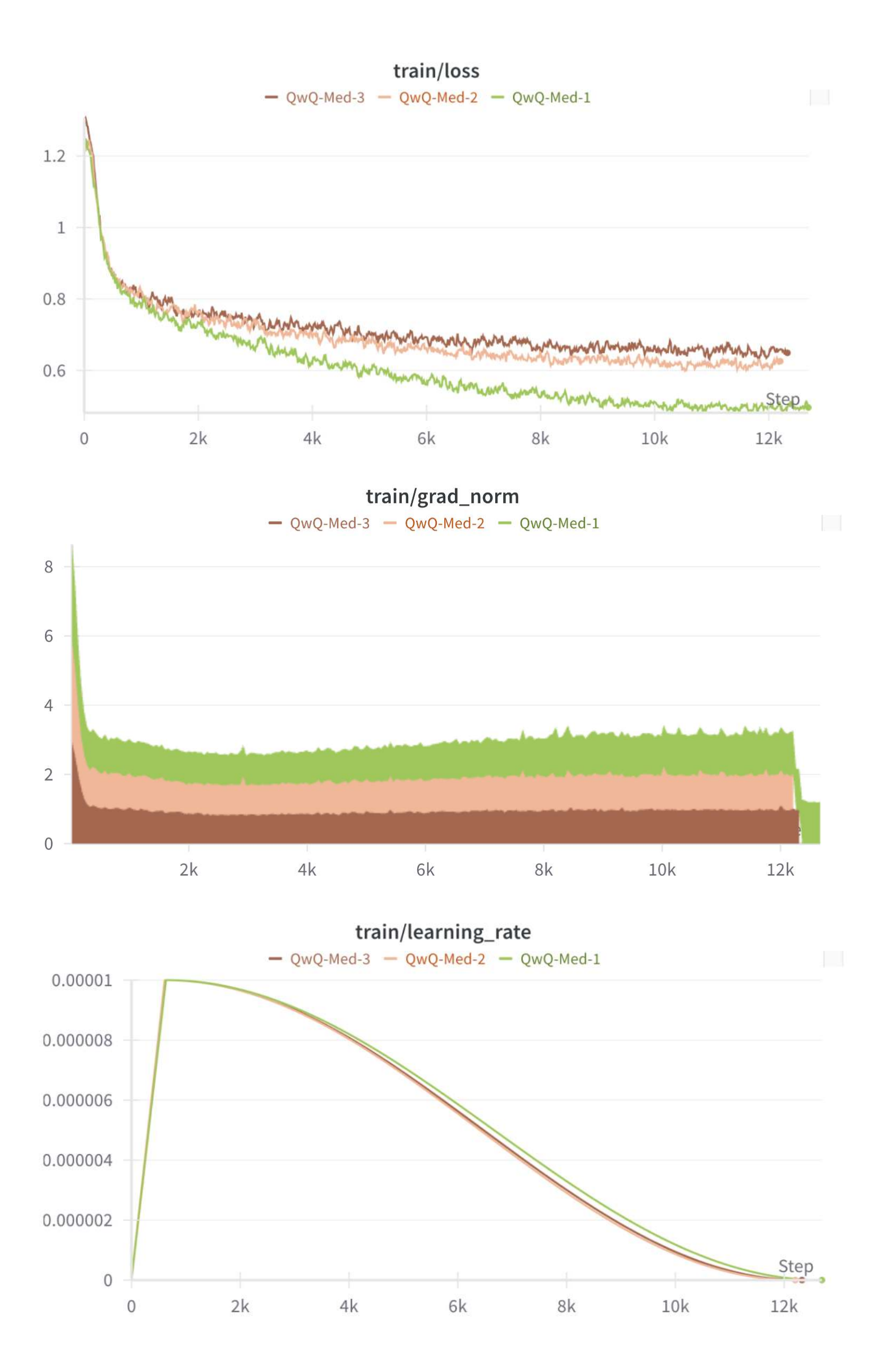}
    \caption{Training run details of our curriculum-tuned models.}
    \label{fig:training_setup}
\end{figure}

We fine-tuned the off-the-shelf QwQ-32B reasoning model \cite{yang2025qwq32b}, which demonstrates strong performance across several medical QA benchmarks and exhibits broad understanding of medical concepts. During supervised fine-tuning (SFT), we injected the thinking trace of the model between \texttt{<think>} and \texttt{</think>} delimiters, followed by the final answer. The next-token prediction objective was applied only to the reasoning trace and the answer. We employed LoRA \cite{Hu2021LoRA:Models} with rank $=16$ and $\alpha=16$, using a batch size of 16 with gradient accumulation every two steps across $8\times$H100 GPUs. The learning rate followed a cosine schedule with a peak of $1\text{e}{-5}$ and linear warmup over the first $5$\% of training steps. We used the Adam optimizer with $\beta_1 = 0.9$, $\beta_2 = 0.95$, and a weight decay of $1\text{e}{-4}$. We ran SFT for 8 epochs on the full dataset. For models trained on subsets, we adjusted the number of epochs to match the total training FLOPs of the full-scale setting. Fig.~\ref{fig:training_setup} visualizes the training dynamics of our fine-tuned models.

\clearpage
\section{Additional Examples: Qualitative Comparison on ICD-Bench}
\label{app:qualitative}
Examples~\ref{comp_1},~\ref{comp_2}, and ~\ref{comp_3} qualitatively compare the outputs of the base QwQ model and our QwQ-Med-3 model. The base model tends to think longer and often in repetitive loops. These examples also illustrate the reasoning-recall gap, where the base model is successful in recalling the entities along the KG path but is often unsure or unable to compose them into coherent reasoning, leading to incorrect conclusions. On the other hand, our curriculum-tuned models demonstrate the characteristic of structured step-by-step thinking traces derived from KG paths, and decisively reason with fewer tokens.\\[2pt]
\refstepcounter{boxcounter}
\begin{tcolorbox}[
    enhanced,            
    breakable=true,           
    colback=pink!20,
    colframe=black,
    title=Example~\theboxcounter: QwQ-32B-Base vs. QwQ-Med-3 Response,
    width=\textwidth,
    label=comp_1
]
\textbf{KG Path from Diseases of the Digestive System:}\\[2pt]Juvenile Colonic Polyposis$\xrightarrow{\text{maybe allelic with}}$Hereditary haemorrhagic telangiectasia$\xrightarrow{\text{maybe causes}}$ Arteriovenous malformation$\xrightarrow{\text{maybe causes}}$Intracranial arteriovenous malformation\\[3pt]
\textbf{Question:}\\[3pt]
 A 14-year-old male presents with recurrent episodes of painless rectal bleeding and iron deficiency anemia. Colonoscopy reveals numerous hamartomatous polyps throughout the colon. Further investigation reveals scattered telangiectasias on his lips and nasal mucosa. Which of the following neurovascular abnormalities should be considered in this patient, given the constellation of findings? 
\\[3pt]
\textbf{Options:}\\[3pt]
A. Moyamoya disease\quad  B. Cerebral amyloid angiopathy \quad \textbf{C. Intracranial arteriovenous malformation} \quad D. Cavernous malformations\\[3pt]
\textbf{QwQ-Base-32B Response (1923 thinking tokens, incorrect answer):}\\[2pt]
\texttt{<think>} Okay, let's tackle this question. The patient is a 14-year-old male with painless rectal bleeding, iron deficiency anemia, and colonoscopy showing hamartomatous polyps throughout the colon. Additionally, there are telangiectasias on his lips and nasal mucosa. The question is asking which neurovascular abnormality to consider here.

First, I need to recall what conditions present with these symptoms. The key points here are the hamartomatous polyps in the colon and the telangiectasias. Hamartomatous polyps make me think of a hamartomatous polyposis syndrome. The most common ones are Peutz-Jeghers syndrome, Juvenile polyposis syndrome, and maybe Neurofibromatosis. But wait, the presence of telangiectasias might point to something else.

Peutz-Jeghers syndrome is characterized by hamartomatous polyps in the GI tract and mucocutaneous pigmentation, usually around the lips, mouth, and genital areas. The pigmentation is often described as "perioral" and "periungual." However, in this case, the telangiectasias are mentioned, which are small dilated blood vessels. Hmm, maybe I need to think of a different condition here.

Wait, another syndrome that includes both polyps and telangiectasias is Hereditary Hemorrhagic Telangiectasia (HHT), also known as Osler-Weber-Rendu disease. But HHT is primarily characterized by telangiectasias and arteriovenous malformations (AVMs) in various organs, including the lungs, liver, and brain. However, do HHT patients get hamartomatous polyps? I think HHT is more about telangiectasias and AVMs, but not necessarily polyps. So maybe that's not the right direction.

Wait, back to the polyps. Let me think again. The question mentions hamartomatous polyps. Juvenile polyps are usually not hereditary unless in Juvenile polyposis syndrome. But the key here is the combination with telangiectasias. Alternatively, there's a condition called PTEN Hamartoma Tumor Syndrome, which includes Cowden syndrome. Cowden syndrome can have multiple hamartomatous polyps and various other manifestations, including vascular malformations. But I'm not sure about telangiectasias specifically.

Alternatively, maybe the patient has a condition that includes both gastrointestinal polyps and neurological issues. Let me think of the possible neurovascular abnormalities listed in the options. The options are Moyamoya, cerebral amyloid angiopathy, intracranial AVM, or cavernous malformations.

Wait, the presence of telangiectasias and polyps might point to a genetic disorder that includes vascular abnormalities. Let me consider Turcot syndrome, but that's more about colorectal cancer and brain tumors like medulloblastoma or glioblastoma, so maybe not. 

Wait, the telangiectasias on lips and nasal mucosa plus hamartomatous polyps. Hmm. Let me think again of Peutz-Jeghers. The pigmentation in Peutz-Jeghers is actually dark spots (due to increased melanin), not telangiectasias. So maybe the telangiectasias here are a clue towards another condition. 

Alternatively, maybe the patient has a condition called Ataxia-Telangiectasia, but that's more about cerebellar ataxia, telangiectasias in eyes, and immunodeficiency. But the GI polyps aren't part of that. 

Wait, the question mentions that the polyps are hamartomatous. Hamartomatous polyps in the colon could be part of a syndrome that includes vascular malformations. Let's think of Neurocutaneous melanosis? No, that's different. 

Wait, the options given are neurovascular abnormalities. Let me look at the options again. 

Option C is intracranial arteriovenous malformation. HHT is associated with AVMs, so if the patient had HHT, then C could be correct. But does HHT present with hamartomatous polyps? I think HHT does have telangiectasias and AVMs but not hamartomatous polyps. 

Wait, maybe the key is that the hamartomatous polyps plus telangiectasias are part of a different syndrome. Let me think again. 

Wait, there's a condition called PTEN Hamartoma Tumor Syndrome (Cowden syndrome) which can have gastrointestinal hamartomas, mucocutaneous lesions (like papillomatous papules, not sure about telangiectasias), and also can have neurological issues. For example, they might have vascular malformations. But the neurovascular abnormalities here could be cavernous malformations? 

Wait, cavernous malformations are vascular lesions composed of blood-filled cavities separated by thin walls. They can occur in various conditions. 

Alternatively, the Peutz-Jeghers syndrome doesn't have cavernous malformations. 

Alternatively, the presence of telangiectasias and polyps might be pointing to a different genetic condition. Let me think of the genetic syndromes associated with both. 

Wait, the options given include cavernous malformations (D). Let me think: some genetic conditions cause both GI polyps and cavernous malformations. 

Alternatively, maybe the hamartomatous polyps and telangiectasias are part of a syndrome where cavernous malformations are common. 

Alternatively, if the telangiectasias are part of HHT, then the neurovascular issue would be intracranial AVM (option C). But is that the case here?

Wait, the question says the patient has "scattered telangiectasias on his lips and nasal mucosa." HHT typically presents with mucocutaneous telangiectasias, especially on the lips, oral mucosa, and nose. So that fits. Also, HHT can have gastrointestinal telangiectasias leading to bleeding. But the question mentions hamartomatous polyps. Wait, do HHT patients get hamartomatous polyps? No, HHT is about telangiectasias and AVMs, not hamartomatous polyps. So that's conflicting. 

Alternatively, maybe the telangiectasias are not due to HHT but another condition. The hamartomatous polyps are in the colon. Let me think again. 

Another possibility is that the patient has a syndrome where both hamartomatous polyps and vascular malformations occur. 

Wait, Cowden syndrome (PTEN mutation) can have hamartomatous intestinal polyps, as well as vascular malformations. In Cowden, they can have vascular malformations such as capillary malformations or cavernous malformations. So maybe cavernous malformations (option D) would be the answer here. 

Alternatively, Neurofibromatosis type 1 can have GI polyps, but I don't recall them being hamartomatous in that context. 

Wait, let me confirm: Cowden syndrome (PTEN-related) is associated with multiple hamartomatous polyps in the GI tract and also has an association with vascular malformations. Specifically, they can have cerebrovascular lesions like cavernous malformations. So the answer might be D, cavernous malformations. 

Alternatively, another thought: the presence of telangiectasias and GI polyps might point to a different condition. Let me think of the possible neurovascular abnormalities. The options are Moyamoya (a specific vascular abnormality), cerebral amyloid angiopathy (usually in older patients), intracranial AVM (HHT), or cavernous malformations (Cowden). 

Since the patient is 14, cerebral amyloid angiopathy is unlikely. Moyamoya is a progressive stenosis of the cerebral arteries, more common in children but not typically linked to these GI findings. 

So between C and D. If the patient has HHT, then the AVM (C) would be the answer, but since the polyps are hamartomatous, which HHT doesn't have, then maybe the correct syndrome is Cowden, leading to cavernous malformations. 

Alternatively, perhaps the telangiectasias are separate from the polyps, but the combination is pointing to Cowden. 

Another angle: the question states "hamartomatous polyps throughout the colon" and "telangiectasias on lips and nasal mucosa". The combination of hamartomas and telangiectasias might be pointing to Cowden syndrome, which includes both. Since Cowden can have neurological issues including vascular malformations (like cavernous), then D is the answer. 

Alternatively, if the telangiectasias are part of HHT, but the polyps are a coincidence? Unlikely, since the question is about the constellation of findings. 

Therefore, the most likely answer is D, Cavernous malformations. \texttt{</think>} Final Answer: D\\[2pt]
\textbf{QwQ-Med-3 Response (820 thinking tokens, correct answer):}\\[2pt]
\texttt{<think>} Okay, let's break this down. 
\begin{enumerate}[leftmargin=1.2em]

\item \textbf{Identify the Syndrome:} First, look at the patient's presentation. We have a 14-year-old male with recurrent painless rectal bleeding (often bright red blood per rectum, or `painless hematochezia') and iron deficiency anemia. The colonoscopy confirms numerous hamartomatous polyps throughout the colon. Additionally, he has telangiectasias (dilated small blood vessels) on his lips and nasal mucosa. This specific combination of symptoms – colonic hamartomatous polyps, rectal bleeding, and mucocutaneous telangiectasias – is highly suggestive of \textbf{Hereditary Hemorrhagic Telangiectasia (HHT), also known as Osler-Weber-Rendu syndrome}.

\item \textbf{Understand HHT:} HHT is an autosomal dominant disorder affecting the vascular system. The classic features are:
\begin{itemize}
    \item \textbf{A:} Epistaxis (nosebleeds) – often severe and recurrent.
    \item \textbf{B:} Telangiectasias (on skin, lips, oral/nasal mucosa).
    \item \textbf{C:} Gastrointestinal bleeding (from telangiectasias or sometimes larger arteriovenous malformations).
    \item \textbf{D:} Pulmonary arteriovenous malformations (PAVMs).
\end{itemize}
Other common features include liver AVMs and \textbf{cerebral vascular malformations}.

\item \textbf{Connect HHT to Neurovascular Abnormalities:} The question specifically asks about neurovascular abnormalities associated with this constellation of findings (which we've identified as likely HHT). HHT is known to cause various vascular malformations throughout the body, including the brain. The most common type of brain vascular malformation seen in HHT is an \textbf{intracranial arteriovenous malformation (AVM)}. These are abnormal connections between arteries and veins in the brain, bypassing the normal capillary system. They can cause symptoms like headaches, seizures, focal neurological deficits, or more seriously, intracranial hemorrhage (bleeding into the brain).

\item \textbf{Evaluate the Options:}
\begin{itemize}
    \item \textbf{A. Moyamoya disease:} This is a progressive narrowing (stenosis) and occlusion of the internal carotid arteries and their main branches, leading to the formation of a network of tiny collateral vessels (the "puff of smoke" or \emph{moyamoya} appearance). While it causes strokes and TIAs, it's a distinct entity not directly linked to HHT.

    \item \textbf{B. Cerebral amyloid angiopathy (CAA):} This involves amyloid protein deposition in the walls of cerebral arteries, primarily affecting older adults (usually >60 years) and causing lobar hemorrhages. It's not associated with HHT.

    \item \textbf{C. Intracranial arteriovenous malformation (AVM):} This is a direct match for a known complication of HHT. The abnormal blood vessel formation characteristic of HHT frequently involves the brain, leading to AVMs. Given the patient's likely HHT diagnosis based on the colonic polyps and mucocutaneous telangiectasias, this is the most probable associated neurovascular abnormality.

    \item \textbf{D. Cavernous malformations:} These are collections of abnormal, thin-walled blood vessels. While they can occur in the brain and cause neurological symptoms or bleeding, they are not specifically characteristic of HHT like AVMs are.
\end{itemize}

\item \textbf{Conclusion:} Based on the patient's presentation (colonic hamartomatous polyps, mucocutaneous telangiectasias) which strongly points towards HHT, the most likely associated neurovascular abnormality to consider is an intracranial arteriovenous malformation (AVM). Therefore, option \textbf{C} is the correct answer.
\end{enumerate}
 \texttt{</think>} Final Answer: \textbf{C}

\end{tcolorbox}

\refstepcounter{boxcounter}
\begin{tcolorbox}[
    enhanced,            
    breakable=true,           
    colback=pink!20,
    colframe=black,
    title=Example~\theboxcounter: QwQ-32B-Base vs. QwQ-Med-3 Response,
    width=\textwidth,
    label=comp_2
]
\textbf{KG Path from Mental, Behavioral, and Neurodevelopmental Disorders:}\\[2pt]Fatal familial insomnia $\xrightarrow{\text{maybe cause}} $Ataxia$ \xleftarrow{\text{maybe cause}}$MECR-related neurologic disorder$\xrightarrow{\text{maybe cause}}$ Choreoathetosis\\[3pt]
\textbf{Question:}\\[3pt]
  A 55-year-old male presents with a rapidly progressive neurological decline over the past 6 months. His wife reports increasing insomnia, characterized by fragmented sleep with frequent, agitated awakenings. Neurological examination reveals dysarthria, myoclonus, and difficulty with gait. Brain MRI is unremarkable. Polysomnography demonstrates a significant reduction in slow-wave sleep and REM sleep. CSF analysis is normal except for slightly elevated levels of total protein. Genetic testing reveals a mutation in the PRNP gene. As his condition progresses, what additional neurological manifestation is MOST likely to develop given the underlying etiology?
\\[3pt]
\textbf{Options:}\\[3pt]
A.  Lower motor neuron weakness with fasciculations.\quad B.  Prominent resting tremor and rigidity. \quad \textbf{C. Involuntary, writhing movements of the limbs and trunk.} \quad D. Loss of vibratory sensation and proprioception in the lower extremities.\\[3pt]
\textbf{QwQ-Base-32B Response (7160 thinking tokens, incorrect answer):}\\[2pt]
\texttt{<think>} Okay, let's tackle this question step by step. The patient is a 55-year-old male with a 6-month history of rapidly progressive neurological decline. His symptoms include insomnia with fragmented sleep and agitation, dysarthria, myoclonus, and gait difficulty. The MRI of the brain is normal, which is a bit confusing because neurological issues often show up there, but maybe not in this case. The polysomnography shows reduced slow-wave and REM sleep. CSF is normal except for slightly high protein. Genetic testing found a PRNP mutation.

Hmm, PRNP gene mutations are associated with prion diseases, like Creutzfeldt-Jakob disease (CJD). I remember that prion diseases are transmissible spongiform encephalopathies. The key features here are the rapidly progressive dementia, myoclonus, and the sleep pattern changes. The sleep issues with reduced slow-wave and REM sleep fit with what I know about CJD. Also, MRI might not show changes early on, but sometimes in CJD, specific MRI findings like high signals in the cortex or basal ganglia can be seen, but the question says it's unremarkable. Maybe the MRI was done early in the disease course.

Looking at the options:

A. Lower motor neuron weakness with fasciculations. Lower motor neuron issues would suggest something like ALS, but prion diseases typically affect upper motor neurons more. ALS has both upper and lower, but the PRNP mutation points away from that.

B. Prominent resting tremor and rigidity. Resting tremor and rigidity are Parkinsonian symptoms. Parkinson's is another possibility, but the rapid progression and the genetic link to PRNP don't fit. Plus, CJD usually doesn't present with Parkinsonian features as a main symptom. The movement disorders in prion diseases are more like myoclonus or ataxia.

C. Involuntary, writhing movements. Writhing movements, like athetosis or chorea, are more typical of basal ganglia disorders, like Huntington's. But Huntington's has a different genetic mutation (HTT gene) and usually a slower progression. However, some prion diseases might have movement issues. Wait, but prion diseases can have ataxia, myoclonus, maybe some dystonia, but writhing movements? Not sure. Maybe not the most likely here.

D. Loss of vibratory sensation and proprioception. That sounds like sensory ataxia, which is seen in conditions like vitamin B12 deficiency, tabes dorsalis, or hereditary sensory neuropathies. In prion diseases, sensory deficits might occur, but more commonly, the cerebellar and cognitive symptoms are prominent. However, some forms of prion disease, like Gerstmann-Sträussler-Scheinker syndrome, can present with ataxia and maybe sensory issues. But the question mentions the PRNP gene mutation, which is more commonly associated with CJD. Also, the patient's initial presentation includes myoclonus and the sleep changes, which are classic for CJD. However, the question is asking what is "MOST likely to develop as the condition progresses".

Wait, in CJD, progression leads to more severe dementia, myoclonus, visual symptoms, maybe akinetic mutism. Sensory loss is less common. But let me think again. The options given: option D is about loss of vibratory and proprioception in lower extremities. That's more of a sensory neuropathy or dorsal column issue. In prion diseases, like in sporadic CJD, the main issues are cognitive, myoclonus, ataxia, visual problems. Maybe the sensory loss isn't a primary feature. Hmm, I'm going back and forth. Let me think of the options again. ............Given the PRNP mutation and the possibility of spinal cord involvement (even in rapidly progressive cases), the most likely additional manifestation is \textbf{D}\texttt{</think>} Final Answer: D\\[2pt]
\textbf{QwQ-Med-3 Response  (1047 thinking tokens, correct answer):}\\[2pt]
\texttt{<think>} Okay, let's break this down.
\begin{enumerate}[leftmargin=1.2em]

\item \textbf{Identify the Diagnosis:} First off, we need to figure out what's going on with this patient. We have a 55-year-old male with a pretty rapid decline over just 6 months. The key features are the neurological problems like trouble speaking (dysarthria), jerky movements (myoclonus), and gait difficulty, plus the really specific sleep disturbance – insomnia with lots of broken sleep and agitation. The brain MRI is clear, which is important because it helps rule out things like strokes or tumors. The CSF is mostly normal except for a tiny bit of protein, which isn't very specific. The real clincher here is the genetic test showing a mutation in the \emph{PRNP} gene.

\item \textbf{What does the \emph{PRNP} gene mutation mean?} This gene codes for the prion protein (PrP). Mutations in \emph{PRNP} cause a group of diseases called prion diseases, which are fatal neurodegenerative disorders. The most common form you might know is Creutzfeldt-Jakob disease (CJD). However, different \emph{PRNP} mutations can cause different clinical pictures, often referred to as familial prion diseases or familial CJD.

\item \textbf{Connecting Symptoms to Prion Disease:} The patient's symptoms fit well within the spectrum of prion diseases.
\begin{itemize}
    \item \textbf{Rapid Neurological Decline:} This is a hallmark of prion diseases.
    \item \textbf{Myoclonus:} This is extremely common in prion diseases, often being a key feature.
    \item \textbf{Dysarthria and Gait Difficulty:} These are also typical neurological signs.
    \item \textbf{Sleep Disturbance:} The description of insomnia with fragmented, agitated sleep is characteristic of a specific type of prion disease called Fatal Familial Insomnia (FFI). FFI is caused by certain specific mutations in the \emph{PRNP} gene. The polysomnography results (reduced slow-wave and REM sleep) further support this type of sleep pathology seen in prion diseases.
    \item \textbf{Unremarkable MRI:} While some changes can eventually be seen, early in the course of many prion diseases, the MRI can be normal.
\end{itemize}

\item \textbf{Evaluating the Options:} The question asks what \emph{additional} neurological manifestation is \emph{most likely} to develop as his condition progresses. We already know he has myoclonus and dysarthria. Let's look at the choices in the context of prion disease progression:

\begin{itemize}
    \item \textbf{A. Lower motor neuron weakness with fasciculations:} This points towards motor neuron disease like ALS. While some rare overlap syndromes exist, isolated lower motor neuron signs aren't a typical primary feature of prion diseases.

    \item \textbf{B. Prominent resting tremor and rigidity:} This sounds like Parkinsonism. Prion diseases can sometimes have Parkinsonian features (like slowness, stiffness), but a \emph{prominent resting tremor} isn't characteristic.

    \item \textbf{C. Involuntary, writhing movements of the limbs and trunk:} These are called \textbf{choreoathetoid movements}. This type of movement disorder is a known feature that can develop in various forms of prion disease as the illness progresses and affects different brain areas, particularly the basal ganglia. Given the underlying diagnosis of a prion disease (\emph{PRNP} mutation), developing choreoathetosis is a plausible and relatively common additional neurological sign.

    \item \textbf{D. Loss of vibratory sensation and proprioception in the lower extremities:} This points to sensory ataxia, often due to a problem in the posterior columns of the spinal cord. While prion diseases can affect sensory pathways (causing sensory loss) or cause spinocerebellar degeneration, this specific pattern isn't as classically associated with the progression of prion diseases as choreoathetosis is.
\end{itemize}

\item \textbf{Conclusion:} Based on the underlying diagnosis of a prion disease (\emph{PRNP} mutation) and the known spectrum of neurological signs that can develop, involuntary writhing movements (choreoathetosis) are a well-recognized additional feature. Therefore, this is the most likely symptom among the choices to appear or become more prominent as the disease progresses.

\end{enumerate}
\texttt{</think>} Final Answer: \textbf{C}

\end{tcolorbox}

\refstepcounter{boxcounter}
\begin{tcolorbox}[
    enhanced,            
    breakable=true,           
    colback=pink!20,
    colframe=black,
    title=Example~\theboxcounter: QwQ-32B-Base vs. QwQ-Med-3 Response,
    width=\textwidth,
    label=comp_3
]
\textbf{KG Path from Neoplasms:}\\[2pt]Adrenal Cortical Carcinoma $\xleftarrow{\text{may treat}} $Mitotane$ \xrightarrow{\text{maybe cause}}$Hypercholesterolaemia\\[3pt]
\textbf{Question:}\\[3pt]
 A 45-year-old female presents with virilization, new-onset diabetes mellitus, and a rapidly growing abdominal mass discovered on imaging. Hormonal evaluation reveals elevated cortisol, androgens, and estrogen levels. She is diagnosed with advanced stage adrenal cortical carcinoma and initiated on a standard treatment regimen. Six months later, follow-up labs show stable disease and controlled hormone levels. However, her lipid panel reveals a significant increase in total cholesterol and LDL cholesterol despite adhering to a low-fat diet. Which of the following is the MOST likely contributing factor to the patient's current hypercholesterolemia? 

\textbf{Options:}\\[3pt]
A. The direct effect of elevated cortisol on hepatic lipid metabolism.\quad B. Prominent resting tremor and rigidity. \quad \textbf{C. A side effect of the medication used to treat the adrenal cortical carcinoma.} \quad D. Increased dietary intake of saturated fats secondary to improved appetite and overall well-being.
\\[3pt]
\textbf{QwQ-32B-Base Response (3672 thinking tokens, incorrect answer): }\\[2pt]
\texttt{<think>}
Okay, let's see. The question is about a 45-year-old female with adrenal cortical carcinoma presenting with virilization, new diabetes, and an abdominal mass. Her hormone levels are elevated in cortisol, androgens, and estrogens. After treatment, her disease is stable, but now she has high cholesterol and LDL despite a low-fat diet. The options are about why her cholesterol is up now.

First, I need to recall what adrenal cortical carcinoma does. These tumors can produce various hormones, like cortisol, androgens, estrogens, etc. Since her hormone levels were initially elevated, but now they're controlled (stable disease and hormone levels controlled), maybe the tumor isn't the direct cause now. But the question says the hypercholesterolemia appeared after treatment. Wait, the follow-up labs show stable disease and controlled hormones, so the tumor isn't actively secreting those hormones anymore? Or maybe they're controlled with treatment?

The cholesterol increase is despite a low-fat diet. So, the cause might be something else besides diet. Let's look at the options.

Option A: Elevated cortisol affecting hepatic lipid metabolism. But the question says her hormone levels are now controlled. If cortisol is under control, then maybe this isn't the cause now. Unless the treatment isn't fully controlling cortisol yet, but the question states "controlled hormone levels," so maybe A is not the answer.

Option B: Tumor-induced ectopic production of apolipoprotein B. ApoB is a key protein in LDL particles. If the tumor is producing ApoB, that could lead to more LDL. However, the tumor is now stable and maybe not as active. But the question says "tumor-induced ectopic" – so maybe the tumor is producing this even when it's stable? Not sure. Alternatively, maybe the tumor's previous effects or other factors.

Option C: Side effect of medication. What's the standard treatment for adrenal cortical carcinoma? Let me think. Treatment might include mitotane, a drug used for adrenocortical carcinoma. Mitotane is an adrenal enzyme inhibitor, but does it affect cholesterol? Or maybe other meds like steroids? Wait, but she has hypercortisolism, so maybe they are using ketoconazole to suppress cortisol. Ketoconazole is an antifungal but also inhibits steroid synthesis. However, some treatments for cancer might include other drugs. Alternatively, if they're using glucocorticoids as part of treatment, but that might not be the case here. Alternatively, maybe statins? Wait, but she's on a low-fat diet, but the question is why her cholesterol is high. If the meds are causing it, like some immunosuppressants or other drugs, but I need to think of common meds for adrenal carcinoma that can raise cholesterol. Maybe mitotane? Not sure. Alternatively, maybe the treatment includes drugs that affect lipid metabolism. Hmm.

Option D: Increased dietary intake due to improved appetite. The question says she's adhering to a low-fat diet, so this is less likely. The answer probably isn't D.

Back to option B. Apolipoprotein B is made in the liver and is part of VLDL and LDL. If the tumor is producing ApoB ectopically, that could increase lipoproteins. But adrenal cortical carcinoma -- do they produce ApoB? I'm not sure. Alternatively, maybe the tumor's androgen or estrogen production could influence lipid levels. But the hormone levels are now controlled, so that's not it. Wait, but maybe the tumor itself, even if stable, is producing something else?

Alternatively, maybe the standard treatment includes a drug that causes hyperlipidemia. Let's think: Mitotane can have side effects like fatigue, nausea, vomiting, but does it affect cholesterol? I don't recall. Let me think about other drugs. Chemotherapy agents might not directly cause high cholesterol. Alternatively, if the patient is on glucocorticoids, but the question states that her cortisol levels are elevated initially but now controlled, so maybe they are suppressing her own cortisol, hence not giving extra steroids. Alternatively, maybe the treatment includes ketoconazole which can have effects on cholesterol? Wait, ketoconazole inhibits steroid synthesis, but also might interfere with cholesterol synthesis? Not sure. Alternatively, maybe the hypercholesterolemia is a paraneoplastic effect even with stable disease. But the tumor is stable, so perhaps it's producing something else.

Alternatively, the tumor might be producing lipids or hormones that stimulate the liver to make more cholesterol. But the androgens and estrogens were controlled. Wait, adrenal cortical carcinoma can sometimes produce various steroids. If the tumor is still making some hormones, even if levels are controlled, maybe through treatment, but perhaps the treatment isn't fully suppressing all hormone production. Wait, the question says "controlled hormone levels," so probably not.

Alternatively, the tumor might be producing apolipoprotein B directly as an ectopic production. 
...............If I can't recall mitotane causing hypercholesterolemia, then B is the better option. Since the tumor could produce apolipoprotein B as an ectopic product, leading to increased LDL. Therefore, the most likely answer is B.

\texttt{</think>} Final Answer: B\\[-2pt]

\textbf{QwQ-Med-3 Response (929 thinking tokens, correct answer):}\\[2pt]
\texttt{</think>} Okay, let's break down this question about the patient with adrenal cortical carcinoma (ACC) and hypercholesterolemia.

\begin{enumerate}[leftmargin=1.2em]

\item \textbf{Understand the Patient's Situation:} We have a 45-year-old female diagnosed with advanced ACC. This is a rare cancer of the adrenal cortex. Key things to note are her initial presentation (virilization, new-onset diabetes, high cortisol, androgens, estrogens) and the treatment course. She was treated, and her disease is stable, and importantly, her hormone levels (cortisol, androgens, estrogens) are now controlled. However, her lipid panel shows high cholesterol and LDL cholesterol, even though she's on a low-fat diet. We need to figure out the most likely reason \emph{why} her cholesterol is high \emph{now}.

\item \textbf{Analyze the Options:}

\begin{itemize}
    \item \textbf{A. The direct effect of elevated cortisol on hepatic lipid metabolism:} This is a very plausible mechanism. Chronic elevation of cortisol (a glucocorticoid) is well-known to mess with liver metabolism. Cortisol increases the production of Very Low-Density Lipoproteins (VLDL) by the liver. VLDL is rich in triglycerides, but as it travels through the blood, its triglycerides are removed, and it gets converted into LDL (the "bad" cholesterol). So, more VLDL means higher LDL. Furthermore, cortisol can also decrease the number of LDL receptors on cells, meaning the body can't clear LDL from the blood efficiently. This is often seen in conditions like Cushing's syndrome, which is characterized by excess cortisol. \emph{However}, the case explicitly states her hormone levels are now \emph{controlled}. While controlling a severe hypercortisolism might take time or might not fully normalize metabolic effects immediately, we \emph{can't} assume her cortisol is still significantly elevated enough to be the \emph{primary} driver of this new hypercholesterolemia, especially since she was treated and labs show controlled levels.

    \item \textbf{B. Tumor-induced ectopic production of apolipoprotein B:} Apolipoprotein B (ApoB) is a key protein component of VLDL and LDL. If the tumor itself were somehow producing ApoB ectopically (in an abnormal place), it could theoretically lead to increased VLDL/LDL. However, this is a highly specific and unusual mechanism. While some tumors can cause paraneoplastic syndromes (abnormal conditions caused by the tumor's presence, not its direct invasion), significant ectopic ApoB production leading to hypercholesterolemia isn't a recognized or common feature of ACC.

    \item \textbf{C. A side effect of the medication used to treat the adrenal cortical carcinoma:} This is a very strong possibility. Advanced ACC is often treated with chemotherapy, and a common first-line agent is Mitotane (o,p'-DDD). Mitotane is known to have various side effects, including metabolic ones. Importantly, hypercholesterolemia (increased cholesterol and LDL) is listed as a potential adverse effect of Mitotane. Since the patient was diagnosed, started treatment (likely including Mitotane), and developed hypercholesterolemia \emph{after} the treatment initiation (and her other hormone levels became controlled), a medication side effect becomes a highly likely explanation. The timing fits perfectly.

    \item \textbf{D. Increased dietary intake of saturated fats secondary to improved appetite and overall well-being:} While it's true that dietary saturated fats contribute to high cholesterol, the patient is explicitly stated to be adhering to a \emph{low-fat diet}. This makes dietary intake the least likely culprit for a \emph{significant} increase in her lipids.
\end{itemize}

\item \textbf{Synthesize and Conclude:} We've ruled out dietary intake (D) because of the low-fat diet. Ectopic ApoB production (B) is too rare and specific. Elevated cortisol (A) could cause it, but her hormones are stated to be controlled. This leaves the medication side effect (C) as the most probable cause. Given that advanced ACC is often treated with Mitotane, and Mitotane can cause hypercholesterolemia, this aligns best with the clinical scenario presented, especially considering the timeline (treatment initiated, then hypercholesterolemia developed despite controlled hormones).

\vspace{1em}
\textbf{Therefore, the MOST likely contributing factor is a side effect of the medication used to treat the adrenal cortical carcinoma.}

\end{enumerate}

\texttt{</think>} Final Answer: \textbf{C}

\end{tcolorbox}
\clearpage
\section{Additional Details: ICD-Bench Results across Task Difficulty}
\label{app:additional_robustness}

\begin{figure}[!tbh]
    \centering
    \includegraphics[width=\linewidth]{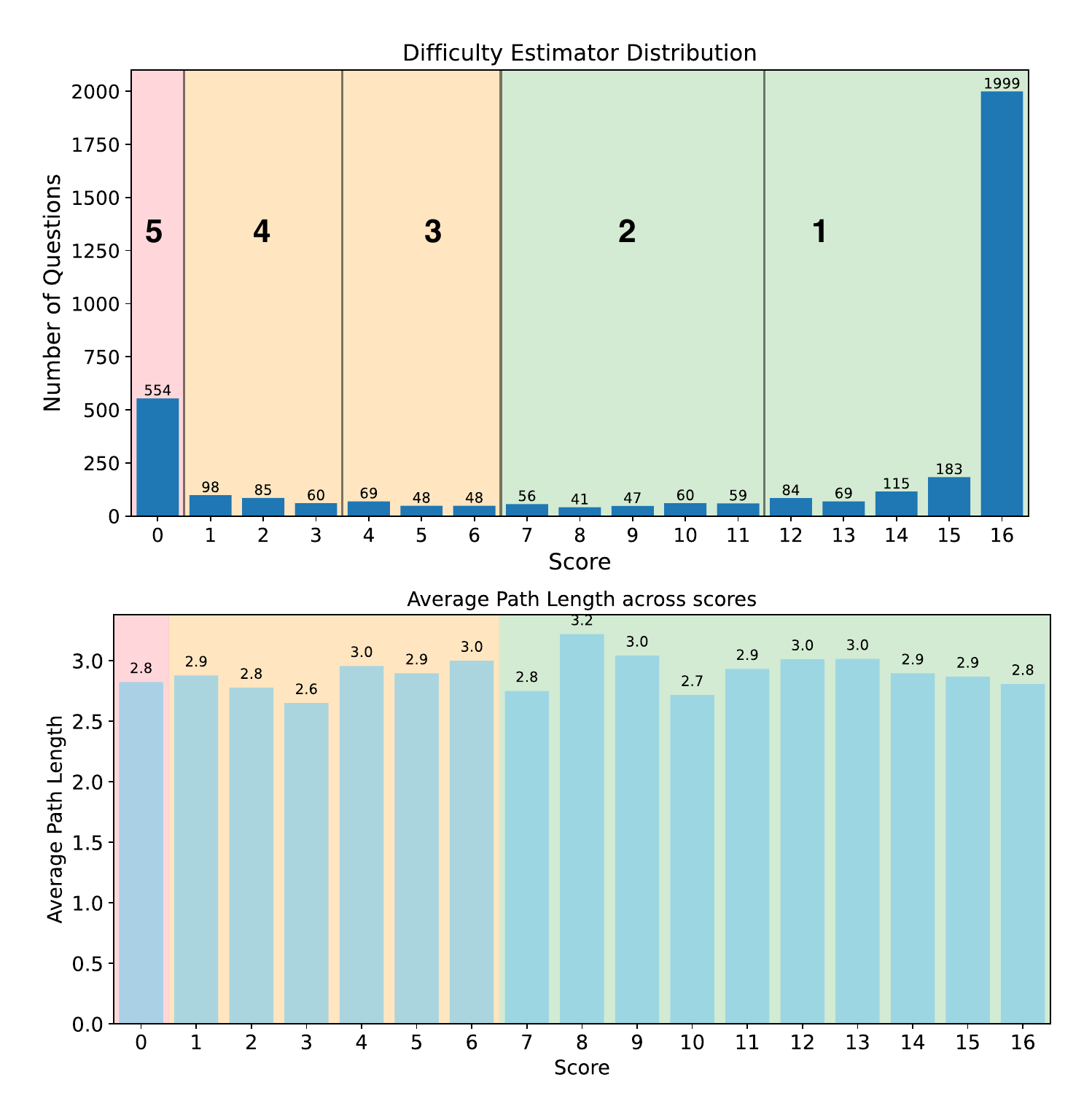}
    \caption{Top: Distribution of $pass@1$ score of the QwQ-Base-32B model over 16 samples, across ICD-Bench questions. Bottom: Mean KG-path length of questions across $pass@1$ scores.}
    \label{fig:task_score}
\end{figure}

\begin{figure}[!hbt]
    \centering
    \includegraphics[width=\linewidth]{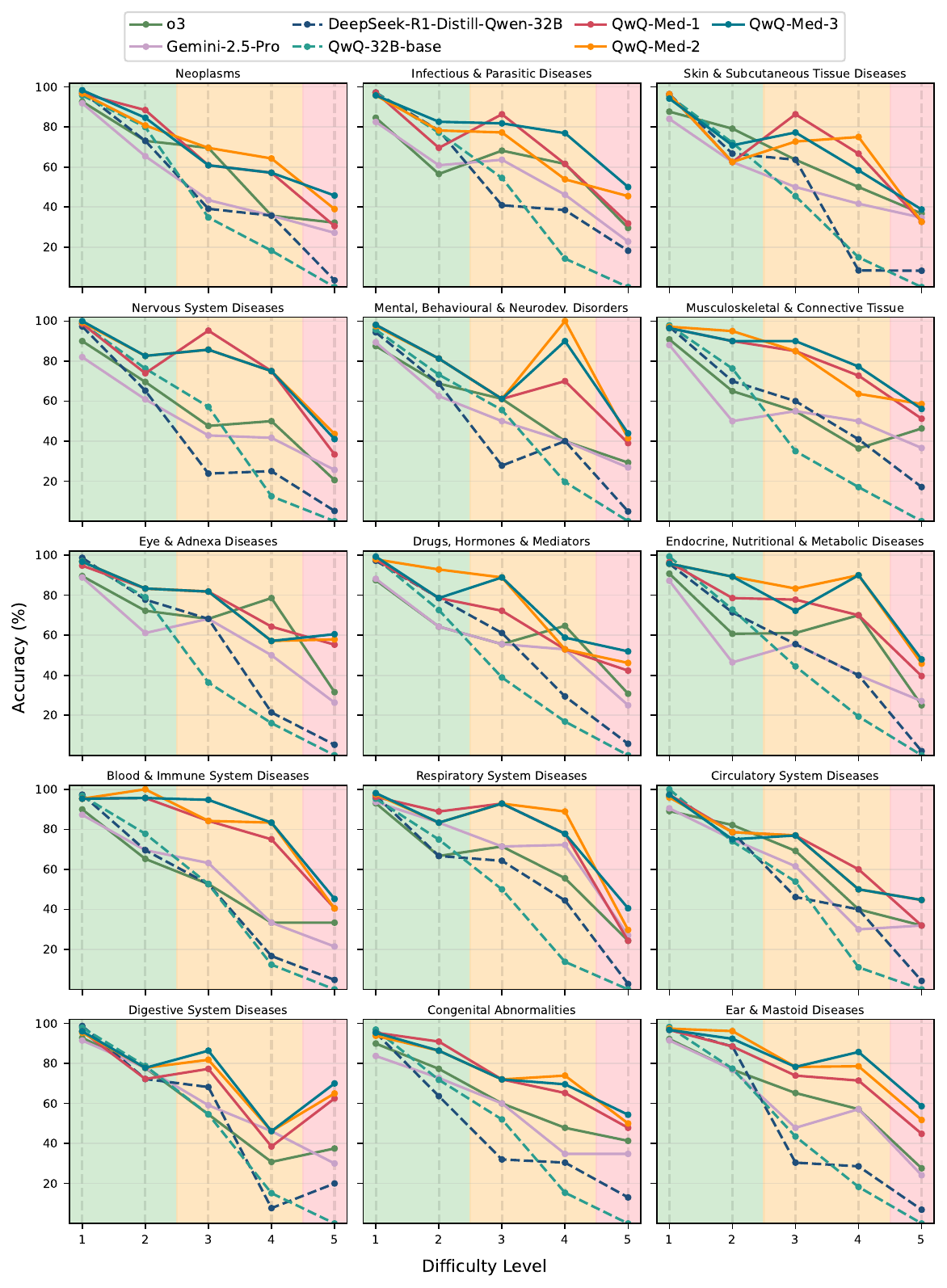}
    \caption{ICD-Bench performance of different models across difficulty levels within each ICD-category.}
    \label{fig:difficulty_wise_category}
\end{figure}

We estimated task difficulty using the $pass@1$ score of the QwQ-32B base model over 16 independent samples. The score distribution across the $ 3,675$ ICD-Bench questions is shown in Fig.~\ref{fig:task_score} (top). We observed a large proportion of full scores, followed by a heavy tail at a zero pass rate. We empirically segmented the questions into five difficulty bins, such that the resulting accuracy of the base model decreases roughly linearly across the bins. In addition, we also quantified the average KG path length of the question for each score in Fig.~\ref{fig:task_score} (bottom) and found no meaningful correlation between the path lengths and the ability of the base model to solve the question. In Fig.~\ref{fig:difficulty_wise_category}, we present the ICD-bench results divided across difficulty levels within each category. We observe that the overall decline in performance with increasing difficulty is consistent across most categories and all models, with curriculum-tuned models exhibiting a widening performance gap at the harder levels.

\clearpage
\section{Additional Details: Medical QA Benchmarks}
\label{app:external}
Prompt~\ref{benchmark_prompt_template} shows the template used to generate responses from both state-of-the-art reasoning and non-reasoning models on existing medical benchmarks. Table~\ref{tab:hyperparams} lists the hyperparameters employed during model inference. Expanded performance results on various MMLU-Med subsets are presented in Table~\ref{tab:medical_models}, including available results of proprietary models, as reported in Ref.~\cite{Yun2025Med-PRM:Rewards}.

\vspace{0.4cm}

\refstepcounter{promptcounter}
\begin{tcolorbox}[
    enhanced,            
    breakable=true,           
    colback=gray!20,
    colframe=black,
    title= Prompt~\thepromptcounter: Template for Medical QA Benchmarks,
    width=\textwidth,
    label=benchmark_prompt_template
]
You are a medical expert presented with an MCQ question. Your final answer should  (A, B, C, or D)

\hspace{1em}<Question>

\hspace{3em}[Clinical Vignette]

\hspace{1em}</Question>

\hspace{1em}<Options>

\hspace{3em}A. [Option]

\hspace{3em}B. [Option]

\hspace{3em}C. [Option]

\hspace{3em}D. [Option]

\hspace{1em}</Options>

\end{tcolorbox}

\begin{table}[h!]
\centering
\scriptsize
\resizebox{\linewidth}{!}{%
\begin{tabular}{lccccc}
\toprule
\textbf{Model} 
  & \textbf{MAX\_NEW\_TOKENS} 
  & \textbf{MAX\_LENGTH} 
  & \textbf{TEMPERATURE} 
  & \texttt{\textbf{TORCH\_DTYPE}} 
  & \textbf{REPETITION\_PENALTY} \\
\midrule
MedGemma  & 2048  & 8192 & 0.6 & bfloat16 & ----  \\
Meerkat   & ----  & 8192 & 0.7 & bfloat16 & 1.2   \\
Deepseek  & ----  & 8192 & 0.6 & bfloat16 & ----  \\
Qwen3     & ----  & 8192 & 0.6 & bfloat16 & 1.2   \\
Sky‑T1    & ----  & 8192 & 0.6 & bfloat16 & ----  \\
QwQ       & ----  & 8192 & 0.6 & bfloat16 & ----  \\
\bottomrule
\end{tabular}%
}
\vspace{1.5em}
\caption{Hyperparameters used for model benchmarking. Unless specified otherwise, default hyperparameters from model providers were used. We used Hugging Face .generate() defaults.}
\label{tab:hyperparams}
\end{table}

\begin{table}[!htb]
  \centering
  \setlength{\tabcolsep}{2.2pt}
  \renewcommand{\arraystretch}{1.05}

  { \scriptsize  
  \begin{tabular}{@{}l@{}*{10}{c}@{}}
    \toprule
    \hdr{Model} &
    \makecell{\hdr{MedQA}\\\hdr{USMLE}} &
    \makecell{\hdr{PubMed}\\\hdr{QA}} &
    \makecell{\hdr{Med}\\\hdr{MCQA}} &
    \makecell{\hdr{MMLU}\\\hdr{(Med-Avg)}} &
    \makecell{\hdr{MMLU}\\\hdr{Clinical}} &
    \makecell{\hdr{MMLU}\\\hdr{Genetics}} &
    \makecell{\hdr{MMLU}\\\hdr{Anatomy}} &
    \makecell{\hdr{MMLU}\\\hdr{Prof Med}} &
    \makecell{\hdr{MMLU}\\\hdr{College Med}} &
    \makecell{\hdr{MMLU}\\\hdr{College Bio}} \\

    \midrule
    \multicolumn{11}{c}{\textbf{Proprietary Models (API only)}} \\[2pt]
    \model{Gemini Flash 2.0}        & 87.51 & --   & 72.60  & 92.01 & --   & --   & --   & --   & --   & --   \\
    \model{GPT-4o-Mini}             & 79.03 & --   & 68.20  & 87.79 & --   & --   & --   & --   & --   & --   \\
    \model{o4-mini}                 & 93.95 & --   & 79.60  & 93.99 & --   & --   & --   & --   & --   & --   \\
    \model{o3-mini}                 & 92.69 & --   & 75.50  & 93.01 & --   & --   & --   & --   & --   & --   \\

    \midrule
    \midrule
    \multicolumn{11}{c}{\textbf{Open-Source Models}} \\[2pt]
    \model{R1-Distill-Qwen (32B)}   & 74.31      & 76.00     & 62.23     & 85.08     & 85.61     & 88.89     & 78.36     & 87.08     & 79.65     & 90.91     \\
    \model{Qwen3 (32B)}             & 64.34      & 65.40     & 55.63     & 69.53     & 73.48     & 72.73     & 57.46     & 69.37     & 65.12     & 79.02     \\
    \model{Sky-T1 (32B)}            & 70.78      & 67.40     & 63.52     & 84.91     & 84.47     & 88.89     & 78.36     & 88.19     & 77.91     & 91.61     \\
    \model{MedGemma (27B)}          & 60.49      & 70.40     & 63.42     & 78.46     & 85.23     & 86.87     & 70.90     & 67.90     & 73.84     & 86.01     \\
    \model{Meerkat (70B)}           & 78.95      & 77.40     & 68.42     & 83.74     & 79.92     & 83.84     & 79.85     & 89.67     & 76.16     & 93.01     \\
    \model{QwQ (32B)}               & \underline{85.62} & 71.00     & 69.26     & 90.46     & 88.26     & 98.00     & 82.09     & 90.77     & 85.47     & 97.20     \\
    \model{QwQ/parallel-scal.}      & \textbf{87.09}    & \underline{78.02} & \underline{71.62} & \underline{91.33} & \underline{89.36} & \underline{98.40} & \underline{83.10} & \underline{93.34} & \underline{85.97} & \underline{97.85} \\

    \midrule
    \multicolumn{11}{c}{\textbf{Our Curriculum-Tuned Model}} \\[2pt]
    \model{QwQ-Med-3 (32B)}         & 82.72      & 76.00     & 71.03     & 90.64     & \underline{89.59} & 98.22     & 81.02     & 92.02     & 85.07     & 96.88     \\
    \model{QwQ-Med-3/parallel-scal.}& 85.39      & \textbf{78.19} & \textbf{73.25} & \textbf{92.90} & \textbf{91.10} & \textbf{98.88} & \textbf{84.84} & \textbf{95.09} & \textbf{89.25} & \textbf{98.01} \\

    \bottomrule
  \end{tabular}
  } 
\vspace{1em}
  \caption{Expanded results from evaluating state-of-the-art models on medical QA benchmarks. Our model, QwQ-Med-3, consistently achieves competitive performance across all benchmarks when compared to similarly sized open-source models. However, it lags behind larger proprietary models, likely due to differences in scale, training compute, and access to private data resources. Best performance on each benchmark is indicated in \textbf{bold}, while second-best is \underline{underlined}, excluding proprietary models.}
  \label{tab:medical_models}
\end{table}

\clearpage

\section{Additional Related Work}
\label{sec:add_related_work}
 
\textbf{Medical Question Answering.} The testbed used for our ICD-Bench is medical question answering, a challenging task that requires models to exhibit reasoning and medical expertise. Traditional systems often rely on access to high-quality external knowledge bases and exhibit limited performance improvements due to the complexity of mapping natural questions to exact answers. The advent of LLMs has led to rapid progress in the field; researchers have introduced benchmarks like MedQA \cite{Jin2020WhatExams}, MedMCQA \cite{Pal2022MedMCQAAnswering}, MMLU \cite{Hendrycks2020MeasuringUnderstanding}, and PubMedQA \cite{jin2019pubmedqa}, derived from medical board exams and clinical topics, to evaluate LLMs on medical reasoning and factual recall. Multiple studies have demonstrated impressive capabilities of LLMs on USMLE and other medical exams, highlighting the benefits of scale and diverse pre-training. Ref.~\cite{Nori2023CapabilitiesProblems} shows GPT-4 exceeding the passing score by more than 20 points on USMLE and outperforming smaller, but fine-tuned, models. Google's Med-PaLM 2 \cite{Google2024SharingBlog} and the MedGemma series \cite{Google2025MedGemmaDevelopers}, specifically tuned on medical data, achieve state-of-the-art performance on benchmarks like MedQA and MedMCQA. Ref.~\cite{Kim2025SmallTextbooks}, with its Meerkat family of models, exhibits the significant potential of small LMs to improve reasoning on medical tasks through careful data curation from verified sources. 

The above results underscore how techniques like knowledge extraction, targeted instruction tuning, and inference-time reasoning collectively contribute to more robust medical QA and diagnostic systems. Yet, challenges regarding integrity and reliability persist; even top models hallucinate and sometimes offer incorrect or unsafe advice. Future work must integrate knowledge-grounded reasoning with rigorous validation to advance domain-specific capabilities while exhibiting the trustworthiness required in critical settings. To address this need, we present a new paradigm for grounding LLMs with verified domain-specific knowledge through KGs and LoRA fine-tuning, thereby enhancing their multi-hop reasoning capabilities. 
    
\textbf{Inference-time Scaling.} Beyond the pre- and post-training, inference-time scaling techniques aim to boost reasoning performance of LLMs during inference. Frontier thinking models, such as OpenAI's o1 and Gemini 2.5 Pro, are known to employ this technique to achieve better results. This involves allocating additional compute or steps at query time to elicit deeper reasoning \cite{muennighoff2025s1simpletesttimescaling}. One paradigm is to prompt the model to produce explicit step-by-step solutions, known as chain-of-thought prompting \cite{wei2023chainofthoughtpromptingelicitsreasoning, Wu2024Chain-of-ThoughCorrection}. More advanced inference-time strategies treat the LLM as an agent that iteratively refines or verifies its own answers \cite{GoogleDeepmind2024AIDeepMind}. So-called ``external'' inference-time scaling methods employ additional models or search procedures alongside the main LLM. Broadly, these techniques can be classified into sequential or parallel inference-time strategies. For example, a best-of-$N$ strategy simply generates $N$ candidate answers in parallel with reasoning and picks the most promising one using a scoring function or a separate verifier model. Other approaches perform a sequential beam search over reasoning steps or use a tree-of-thoughts where the model explores multiple branches of reasoning and a voting or value function identifies the best path \cite{Bi2024Forest-of-Thought:Reasoning, Franceschelli2024CreativeGeneration}. 
    
A recent study by Liu et al. demonstrates that with the right inference-time scaling, a relatively small 1B-parameter model can outperform a 405B-parameter model on complex math problems \cite{Liu2025CanScaling}. Scaling inference-time compute offers a promising direction to improve the capability of LLMs on complex multi-hop reasoning tasks. We employ both parallel and sequential inference-time scaling strategies with our model, demonstrating improvement across ICD-Bench.


\newpage

\end{document}

%% file: math_commands.tex
\usepackage{amsmath,amsfonts,bm}

\def\eqref#1{equation~\ref{#1}}

\def\1{\bm{1}}

\DeclareMathAlphabet{\mathsfit}{\encodingdefault}{\sfdefault}{m}{sl}
\SetMathAlphabet{\mathsfit}{bold}{\encodingdefault}{\sfdefault}{bx}{n}

